\pdfoutput=1
\documentclass[sigconf]{acmart}
\renewcommand\footnotetextcopyrightpermission[1]{}
\setcopyright{none}
\acmConference[Preprint]{Preprint}{2026}{}
\acmYear{2026}
\copyrightyear{2026}
\acmDOI{}
\acmISBN{}

\usepackage{enumitem}
\microtypesetup{expansion=false}
\usepackage{amsmath}
\usepackage{multirow}
\usepackage{makecell}

\usepackage{xspace}
\usepackage{algorithm}
\usepackage{algpseudocode}
\usepackage{tikz}
\usetikzlibrary{positioning,shapes.geometric,arrows.meta,fit,backgrounds,calc}
\newcommand{\itin}{Trip Itinerary\xspace}
\newcommand{\pack}{Packing Checklist\xspace}
\newcommand{\budget}{Travel Budget\xspace}
\newcommand{\safe}{sequential-safe\xspace}
\newcommand{\prone}{coupling-prone\xspace}

\newcommand{\squishlist}{
\begin{list}{$\bullet$}
  { \setlength{\itemsep}{0pt}
     \setlength{\parsep}{0pt}
     \setlength{\topsep}{0pt}
     \setlength{\partopsep}{0pt}
     \setlength{\leftmargin}{1em}
     \setlength{\labelwidth}{0.8em}
     \setlength{\labelsep}{0.2em} } }
\newcommand{\squishend}{
  \end{list}  }

\newcommand{\floatgap}{8pt}
\newcommand{\captiongap}{3pt}
\newcounter{takeaway}
\newcommand{\takeaway}[1]{\par\smallskip\noindent\refstepcounter{takeaway}\fbox{\parbox{\dimexpr\columnwidth-2\fboxsep-2\fboxrule\relax}{\textbf{Takeaway \Roman{takeaway}.} #1}}\par\smallskip}

\tikzset{
  plbox/.style={rounded corners=2pt, draw=black, thick, fill=gray!6, font=\scriptsize, align=center, inner sep=2.5pt, minimum height=5mm, minimum width=11mm},
  plsup/.style={rounded corners=2pt, draw=black, thick, dashed, fill=gray!25, font=\scriptsize, align=center, inner sep=2.5pt, minimum height=6mm, minimum width=12mm},
  plpool/.style={cylinder, shape border rotate=90, aspect=0.28, draw=black, thick, fill=gray!18, font=\scriptsize, align=center, inner sep=1.5pt, minimum height=8mm, minimum width=10mm},
  plgate/.style={diamond, draw=black, thick, fill=gray!12, font=\scriptsize, align=center, inner sep=0pt, aspect=1.7, minimum width=15mm, minimum height=8mm},
  plder/.style={-{Latex[length=1.5mm]}, draw=black, thick, font=\tiny},
  plproj/.style={-{Latex[length=1.5mm]}, draw=black, thick, densely dashed, font=\tiny},
  plclaim/.style={-{Latex[length=1.5mm]}, draw=black, thick, densely dotted, font=\tiny},
  plfeedback/.style={-{Latex[length=1.5mm]}, draw=black, thick, loosely dotted, font=\tiny},
  lbl/.style={font=\tiny, text=black},
}

\begin{document}

\title{When Local Guarantees Do Not Compose: Orchestrating Agentic Workflows under Shared Resource Constraints}

\author{Himel Dev}
\affiliation{
  \institution{529 Tech LLC}
  \city{Renton}
  \country{United States}
}
\email{himel@529-tech.com}

\author{Tanmoy Sen}
\affiliation{
  \institution{University of Wisconsin-Stout}
  \city{Menomonie}
  \country{United States}
}
\email{sent@uwstout.edu}

\author{Sabid Bin Habib Pias}
\affiliation{
  \institution{University at Albany}
  \city{Albany}
  \country{United States}
}
\email{sabhabib@albany.edu}

\author{Madhusudan Basak}
\affiliation{
  \institution{University of Massachusetts Amherst}
  \city{Amherst}
  \country{United States}
}
\email{mbasak@umass.edu}

\author{Joy Ghosh}
\affiliation{
  \institution{529 Tech LLC}
  \city{Renton}
  \country{United States}
}
\email{joy@529-tech.com}

\author{Bashima Islam}
\affiliation{
  \institution{University of Massachusetts Amherst}
  \city{Amherst}
  \country{United States}
}
\email{bashima@umass.edu}

\begin{abstract}
Agentic systems answer a request by composing separately built components, and a guarantee that holds for each component need not hold for their composition when the components draw on a shared, uncertain resource. We study this problem in travel planning, where an itinerary planner, a budget planner and a packing system are each sound alone but together spend one budget. On 2{,}728 requests over 88 trips, running them in sequence breaks a shared limit for 20.7\% of the budgeted requests that have a feasible plan. Frontier language models orchestrating the same capabilities exceed the money limit in 15--30\% of cases, even with a verifier that rechecks plans on the agent's own cost scenarios; certifying their plans on fresh scenarios removes these overruns but leaves a valid plan for only 32--41\% of the requests that have one. We report on an orchestrator that composes the three capabilities unchanged. A slack gate estimates from one joint cost model whether a shared resource is at risk, only the components that draw on such a resource are solved jointly, and every released plan is certified on cost scenarios independent of the gate's sample. The orchestrator releases no plan that an independent re-simulation rejects, returns a plan wherever an exhaustive search over the same components finds one (mean utility gap 0.0067), and needs about half as many component calls as always solving jointly. Building it taught us four lessons: the margin on a shared resource must be computed on the joint total; a plan selected on a sample appears safer on that sample than it is, so that 22\% of releases fail without certification; certification corrects sampling error but not model error; and the check that decides whether to coordinate must itself be cheap.
\end{abstract}

\maketitle

\section{Introduction}
\label{sec:intro}

Agentic systems answer a request by composing specialized components, each built and tested on its own~\cite{yao2023react,shen2023hugginggpt,li2026chainrec,yang2026prta}, and each component can keep its own constraints while the combined answer breaks a limit the components share. Consider a travel assistant built from an itinerary planner, a budget planner and a packing system, two of them running in a production travel app. A traveler asks for four days in Lisbon within \$2,800, planned and packed for. Each component keeps its own constraints, yet together they can describe a trip that costs more than \$2,800, because the attractions, flights and hotel spend the same money and their prices rise and fall together. The problem arises wherever separately built components spend one uncertain resource: queries that each fit their memory grant can together exhaust the memory the database shares among them~\cite{storm2006stmm}, campaigns that each meet their targets can overspend an advertiser's budget~\cite{agarwal2014pacing}, and agents that each finish their subtask can together exceed a token budget~\cite{khan2026tokenbudgets,yang2026bamas}.

The failure is frequent. Running the three components one after another, as a pipeline of the existing systems would, releases a plan that breaks a shared limit for 51\% of our requests, and for 20.7\% of the budgeted requests for which a feasible plan exists. A more capable orchestrator does not remove the problem: frontier language models given the same capabilities as tools and the same budget rule release plans over the money limit in 15--30\% of cases, most of them just over it (§\ref{sec:rq5}).

Two simple remedies fail for opposite reasons. Solving every request as one joint problem keeps the limit but multiplies the components' decision spaces and cost, even where the limit was never at risk. Dividing the limit in advance, or giving each component its own safety margin, keeps the components separate, but correlated costs make any fixed division wrong in one direction or the other.

Whether a limit binds, and which components must then be decided together, is known only at run time, once the request's numbers are in. Two conditions decide it. The first is \emph{binding}: components that share a resource must be decided together only when it is tight enough that deciding them one after another could overrun it. Whether the Lisbon budget binds depends on how tight \$2,800 is for that trip, not on how the request is phrased, so no routing by request type can decide it. The second is \emph{selectivity}: when a resource binds, only the components that claim it, and those they depend on, must be solved together. Packing claims none of the budget, so it can still run on its own.

This paper reports on building and evaluating an orchestrator for \emph{personal logistics}, the connected decisions of what to do, what to pack and what to spend on a trip. It composes three existing capabilities without changing them: an itinerary planner whose schedules are feasible by construction and ranked by a learned preference model~\cite{dev2026feasibleconstruction,dev2026planlearnadapt}, a packing system that enforces bag capacity and safety rules with a constraint solver~\cite{dev2026hardrules}, and a budget planner that chooses bookable options under a chance constraint on total spend~\cite{dev2026budgeteer}. Each capability enters as a typed node whose contract states what it needs, what it guarantees and how much of each shared resource, or \emph{pool}, it claims. Before planning anything, a \emph{slack gate} estimates the request's total spend from one cost model in which flight, hotel and attraction prices move together. If a high quantile of that total fits the cap, the orchestrator plans the itinerary, re-checks the estimate with the chosen attractions, and moves on. If not, it hands the itinerary and the budget plan, and only those, to a joint solver (their \emph{resource closure}) and plans packing afterward. Before release, the plan's overrun probability is certified on cost scenarios independent of the gate's sample. A language model only parses requests and explains answers, because language models alone rarely satisfy every hard constraint of such plans~\cite{dev2026planlearnadapt,dev2026hardrules,xie2024travelplanner}. The same machinery answers ``no plan'' with the smallest change in each dimension that would give one, and replans after a change by re-opening only the closures the change reaches.

We evaluate on 2{,}728 requests over 88 trips, at ten budget levels around each trip's feasibility floor, against always-sequential and always-joint execution, validate-and-repair, fixed budget splits, ablations of the gate and frontier language models orchestrating the same capabilities. An independent validator re-simulates the cost of every released plan, and an exhaustive oracle audits plan quality. Five research questions structure the evaluation: what it takes to keep composed plans within their shared limits (RQ1, \emph{feasibility}); when pools bind and when solving only their closures pays off in time (RQ2, \emph{efficiency}); how much plan quality selective solving gives up (RQ3, \emph{quality}); whether the same machinery can answer ``no plan'' with minimal relaxations and replan after a change (RQ4, \emph{recovery}); and how language models perform as parsers and as orchestrators (RQ5).

The gated system releases no plan the validator rejects (RQ1). Binding follows tightness, not request type: some pool binds for 88\% of requests at 1.1 times the floor and 20\% at 1.5 times (RQ2). On our budget grid the gated system takes the least time per accepted plan of the methods that keep every limit, its saving grows with slow components and with more capabilities, and on loose budgets chosen by users validate-and-repair is as fast. It finds every plan the oracle finds, with a mean utility gap of 0.0067 (RQ3). Relaxations cover every change the oracle confirms, and replanning keeps 0.81 of the itinerary after an attraction closes, against 0.35 from scratch (RQ4). A hybrid parser reads requests almost perfectly, while language-model orchestrators give a valid plan for 61--69\% of plannable requests at one to two orders of magnitude more time (RQ5).

Building the system taught us four lessons. First, the margin on a shared resource must be computed jointly: giving each component its own 90\% bound yields a plan for half as many requests as one joint bound (30\% against 60\%), and treating the cost lines as independent releases plans that overrun in 92\% of cases. Second, a plan chosen as the best on a sample of cost scenarios looks safer on that sample than it is: our first version passed its own checks on every plan, yet an independent re-simulation rejected 22\% of releases. Certifying each plan on a fresh sample removes these overruns at no loss of plans. Third, certification fixes sampling error but not model error: when costs spread 25\% wider than modeled, only 0.7\% of released plans stay within the limit, for the gated system, always-joint and validate-and-repair alike. Fourth, deciding whether to coordinate must itself be cheap: the gate first lost to validate-and-repair in time because it simulated the same plan's cost again and again, and remembering those simulations made it 1.5 times faster. Each effect is known on its own~\cite{artzner1999coherent,smith2006optimizer,luedtke2008sample}; what our system shows is how large they are when existing components are composed, and where in execution each must be handled.

We make the following contributions:
\squishlist
\item \textbf{An orchestrator that composes existing components under shared limits} (§\ref{sec:problem}, §\ref{sec:arch}): typed contracts and pools with one joint chance-constrained margin; execution that gates before it proposes, solves only resource closures jointly and certifies each release on scenarios independent of the gate's sample. The components are used unchanged, two of them from a production app.
\item \textbf{Lessons on what feasibility takes} (§\ref{sec:rq1}): per-component margins, independent cost lines and fixed splits fail; judging a plan on the sample it was chosen on hides overruns, which certification removes; the remaining risk lies in the cost model.
\item \textbf{An evaluation of when coordination pays} (§\ref{sec:rq2}, §\ref{sec:rq3}): binding follows budget tightness, not request type; gating makes 54\% of always-joint's proposals and loses no plan an exhaustive search over the same planners finds.
\item \textbf{Recovery through the same machinery} (§\ref{sec:rq4}): single-dimension relaxations cover every change an oracle confirms, and closure-bounded replanning keeps most of the itinerary in place after an attraction closes.
\item \textbf{A comparison with language models in the loop} (§\ref{sec:rq5}): a hybrid parser reads requests almost perfectly, but frontier models orchestrating the same capabilities with the same budget rule release plans over the money limit, give a valid plan for fewer plannable requests and take far longer.
\squishend

\section{Problem}
\label{sec:problem}
We describe the setting the orchestrator works in, define when a composed plan is feasible, state the problem, and give the two conditions under which components must be decided together.

\subsection{Components and Local Contracts}
\label{sec:capabilities}
The orchestrator composes $K$ existing components, which we also call capabilities; inside the orchestrator each appears as one or more nodes (§\ref{sec:contracts}). Component $k$ takes typed context, such as the destination, dates and travelers, and returns a decision $x_k$ from its \emph{locally feasible set} $\mathcal{F}_k$, the decisions that satisfy its own constraints. Each component is a black box with three operations: it can \emph{propose} decisions for a context, optionally under a cap on what it may claim; \emph{score} a decision with its own utility; and report the decision's \emph{claim} on each resource it shares, as a number or, for uncertain costs, as samples from a cost model. The orchestrator does not change a component's code, objective or search. This is the constraint an industrial system is built under: the components are developed, tested and, in two cases, deployed on their own.

Our three components share a generate, learn and reconcile pattern. \itin~\cite{dev2026feasibleconstruction,dev2026planlearnadapt} produces multi-day schedules under opening hours, travel times and daily time budgets: several planners each return a feasible schedule, and a Bradley--Terry reward model~\cite{bradley1952rank} trained on 2,519 pairwise preferences ranks them, winning 67.8\% of preference comparisons, 11.2 points above the best single planner. \pack~\cite{dev2026hardrules} produces a checklist and bag assignment: a rule engine proposes items that respect safety and regulatory rules, learned models estimate inclusion, and CP-SAT~\cite{perron2023ortools} enforces bag capacity and item dependencies, finding feasible assignments where greedy selection succeeds 28\% of the time. \budget~\cite{dev2026budgeteer} enumerates bookable combinations of flights, lodging, local transport, food and activities, keeps the frontier that trades expected cost against convenience, and chooses the most convenient plan whose cost stays within its cap with probability $\alpha$. \itin{} and \pack{} run in a production travel app. Each guarantee is local: none of the components sees what the others consume.

\subsection{Shared Pools and Global Feasibility}
\label{sec:composition}
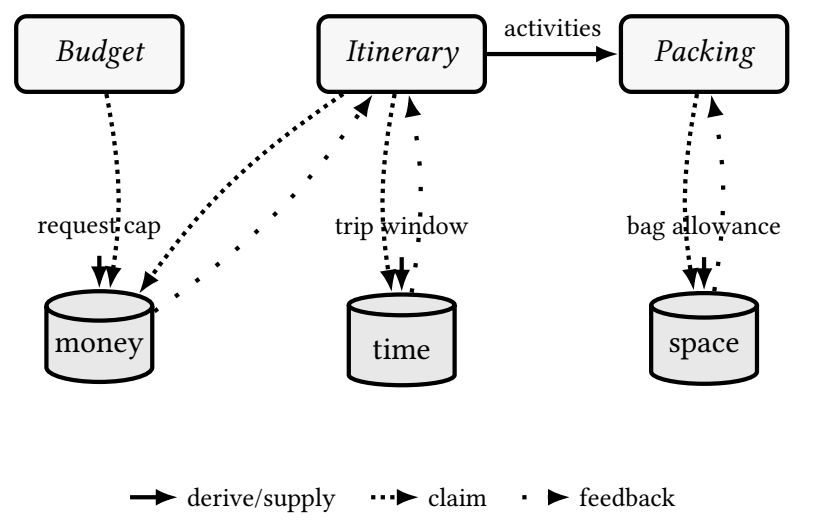
\begin{figure}[t]
\centering
\begin{tikzpicture}
  \node[plbox] (bu) at (-2.0,1.20) {\itshape Budget};
  \node[plbox] (it) at ( 0.0,1.20) {\itshape Itinerary};
  \node[plbox] (pk) at ( 2.0,1.20) {\itshape Packing};
  \node[plpool, minimum height=6mm, minimum width=7mm] (mon) at (-2.0,-0.75) {money};
  \node[plpool, minimum height=6mm, minimum width=7mm] (tim) at ( 0.0,-0.75) {time};
  \node[plpool, minimum height=6mm, minimum width=7mm] (spc) at ( 2.0,-0.75) {space};
  \node[lbl] (mctx) at (-2.0,0.05) {request cap};
  \node[lbl] (tctx) at ( 0.0,0.05) {trip window};
  \node[lbl] (sctx) at ( 2.0,0.05) {bag allowance};
  \draw[plder] (mctx) -- (mon);
  \draw[plder] (tctx) -- (tim);
  \draw[plder] (sctx) -- (spc);
  \draw[plder] (it) -- node[lbl,above]{activities} (pk);
  \draw[plclaim] (it) to[bend right=10] (mon);
  \draw[plfeedback] (mon) to[bend right=10] (it);
  \draw[plclaim] (bu) to[bend left=10] (mon);
  \draw[plclaim] (it) to[bend right=10] (tim);
  \draw[plfeedback] (tim) to[bend right=10] (it);
  \draw[plclaim] (pk) to[bend right=10] (spc);
  \draw[plfeedback] (spc) to[bend right=10] (pk);
  \node[lbl, below=5.5mm of tim.south] {%
    \tikz\draw[plder](0,0)--(3.2mm,0);\ derive/supply \quad
    \tikz\draw[plclaim](0,0)--(3.2mm,0);\ claim \quad
    \tikz\draw[plfeedback](0,0)--(3.2mm,0);\ feedback};
\end{tikzpicture}
\caption{Money is claimed by two capabilities, so its limit is a cross-capability constraint; time and space are each claimed by one. Typed context sets every capacity.}
\Description{Budget, itinerary, and packing capabilities interact through money, time, and space pools. Request context supplies pool capacities, budget and itinerary both claim money, itinerary derives activities for packing, capability-to-pool dotted arrows are claims, and pool-to-capability dotted arrows carry residual allocation feedback.}
\label{fig:coupling}
\end{figure}

Composition can fail when locally feasible decisions spend the same resource, which we call a \emph{pool}. Typed context sets its capacity $B_P$, and components make \emph{claims} on it; what remains limits the decisions that follow (\emph{allocation feedback}). Three pools recur in personal logistics (Figure~\ref{fig:coupling}). \emph{Money} is set by the request's cap and claimed by the budget plan and the itinerary's paid attractions; \emph{time} is set by the travel dates and claimed by the itinerary; \emph{space} is set by the bag allowance and claimed by packing, whose items depend on the itinerary's activities. Money is the one pool that two components claim directly.

Let $c_{kP}(x_k)$ be component $k$'s claim on pool $P$. A composed plan $x=(x_1,\ldots,x_K)$ is \emph{globally feasible} when it belongs to
\begin{equation}
\mathcal{F}_{\mathrm{global}}=\Big\{x\in\textstyle\prod_k\mathcal{F}_k:\ \sum_k c_{kP}(x_k)\le B_P\ \ \forall P\Big\},
\label{eq:global}
\end{equation}
a strict subset of $\prod_k\mathcal{F}_k$ whenever some combination of locally feasible decisions overruns a pool. Time and space claims are known once the decisions are made; money claims are not, since a chosen hotel sets a distribution over its price, and flight, lodging and attraction prices rise and fall together with demand. For such a pool the limit becomes a chance constraint~\cite{charnes1959chance},
\begin{equation}
\Pr\nolimits_M\Big(\textstyle\sum_k c_{kP}(x_k)\le B_P\Big)\ \ge\ \alpha,
\label{eq:chance}
\end{equation}
where $\alpha$ is the required probability of staying within the limit and the claims are random variables under \emph{one joint cost model} $M$ shared by all components. With correlated claims, the $\alpha$-quantile of the total is not the sum of the components' $\alpha$-quantiles, and treating the claims as independent understates the total's spread. Both shortcuts fail in our system, in opposite directions (§\ref{sec:rq1}).

\subsection{Problem Statement}
\label{sec:statement}

Given a request with its context and pool capacities, components reached only through propose, score and claim, a joint cost model $M$, a required probability $\alpha$ and a confidence $1-\delta$, the orchestrator must return a plan or ``no plan''. Conditions 1--3 define when a returned plan is feasible, condition 4 states the objective, and conditions 5--6 are requirements on the orchestrator itself:
\begin{enumerate}[leftmargin=*, itemsep=0pt, parsep=0pt, topsep=2pt]
\item \emph{Local contracts.} Every $x_k\in\mathcal{F}_k$, as the component itself reports.
\item \emph{Deterministic pools.} Every deterministic pool, here time and space, satisfies Eq.~\eqref{eq:global}.
\item \emph{Certified uncertain pools.} For every uncertain pool, Eq.~\eqref{eq:chance} holds with confidence $1-\delta$, established on cost scenarios drawn from $M$ independently of the sample on which the gate judges plans.
\item \emph{Quality.} Among plans that meet 1--3, it prefers higher utility, computed from the components' own scores; we measure the gap to an exhaustive search over the same components (§\ref{sec:rq3}).
\item \emph{Recovery.} When it returns ``no plan'', it offers, for each dimension of the request, the smallest change that gives a plan.
\item \emph{Cost.} It solves components jointly only where 1--3 require it (§\ref{sec:taxonomy}), so a request with loose limits costs about as much as running the components one after another.
\end{enumerate}
In our instance, condition 3 requires the one-sided Clopper--Pearson upper bound on the overrun probability~\cite{clopper1934}, on $n$ fresh scenarios, to be at most $1-\alpha$, with $\alpha=0.9$, $n=2{,}000$ and $\delta=0.05$. The utility weighs the itinerary's reward and the budget plan's convenience equally, and a request's dimensions are its budget, trip length and month. The guarantee in condition 3 is relative to $M$: certification removes the error of judging a plan on the sample it was chosen on, not the error of a model whose prices are wrong. Section~\ref{sec:rq1} measures both with an independent validator. The problem is the same wherever separately built components spend one uncertain resource; only the components, pools and cost model change.

\subsection{When to Solve Jointly}
\label{sec:taxonomy}
Running the components one after another meets conditions 1--3 unless a pool is tight, while solving them all jointly meets them always at the price of multiplying their decision spaces. The orchestrator therefore decides, as it executes, which components to decide together. Let $A$ be the components whose decisions are accepted so far and $U$ those still unresolved.
\squishlist
\item \emph{Binding.} Pool $P$ \emph{binds} when the $\alpha$-quantile of its total claim exceeds $B_P$, with accepted claims conditioned on their decisions and unresolved claims at their \emph{ex-ante estimates}, the claims components report before they decide (§\ref{sec:slack}). Deciding the unresolved claimants one after another could then overrun the pool.
\item \emph{Selectivity.} When $P$ binds, only its \emph{resource closure} $H_P\subseteq U$, the unresolved claimants of $P$ and the unresolved components they depend on, must be decided together.
\squishend
Selective solving saves work only when pools often do not bind or $H_P$ is smaller than $U$, which is an empirical question (§\ref{sec:rq2}). Request types are a natural alternative guide; we treat a type only as a structural prior, \safe or \prone, because binding follows how tight the cap is for each trip (Table~\ref{tab:taxonomy}, §\ref{sec:setup}).

\begin{table}[t]
\centering\footnotesize
\setlength{\tabcolsep}{3pt}
\caption{The structural prior and how often a pool bound in the gated runs, at 1.1 and 1.5 times the trip's cost floor (share of requests; follow-ups: share opening the money closure).}
\label{tab:taxonomy}
\begin{tabular}{@{}lllcc@{}}
\toprule
Type & Example & Prior & $\tau=1.1$ & $\tau=1.5$ \\
\midrule
Atomic & ``plan 3 days in Rome'' & \safe & \multicolumn{2}{c}{(no cap)} \\
Joint budget & ``Rome: tours + food $\le$ \$400'' & \prone & 82\%& 0\%\\
Allocation & ``split \$2,000 three ways'' & \prone & 91\%& 2\%\\
Conflict & ``carry-on ski trip $<$ \$2,500'' & \prone & 92\%& 59\% (bag) \\
What-if & ``swap the museum for a park'' & \safe & 70\%& 22\%\\
Replanning & ``rained out, over budget'' & \prone & 81\%& 19\%\\
Explanation & ``why is this infeasible?'' & \safe & -- & -- \\
\bottomrule
\end{tabular}
\end{table}

\section{Architecture}
\label{sec:arch}

The system separates interpretation from enforcement in four layers. The Language layer maps text to a typed request and renders answers; it is the only layer that calls a language model. The Orchestration layer resolves the capability graph, attaches pools, runs the slack gate, and validates every contract and certifies every uncertain pool before release. The Context layer holds typed facts, commitments and pool state, and the Capability layer holds the typed nodes. Hard-constraint reasoning thus stays in control code and tools, never in free-form generation.

Algorithm~\ref{alg:resolve} gives the execution of one request. The orchestrator resolves the request into a graph of typed nodes and attaches the pools (lines 1--2), then repeats one step until no node is unresolved. It computes the margin of every pool (line 4). If no pool binds, it asks the next node for a proposal and accepts it when no pool binds under it (lines 12--15); if a pool binds, it solves that pool's resource closure jointly (lines 6--10). Certification on fresh scenarios decides acceptance at every step, and before release every contract is validated and every uncertain pool certified once more (lines 22--23). For Lisbon with a loose cap, each node is proposed once and nothing is solved jointly; with a tight cap, money binds before the itinerary is planned, the itinerary and the budget plan are solved together, and packing runs afterward.

\subsection{Typed Contracts and Resolution}
\label{sec:contracts}
Each capability enters the graph as one or more typed \emph{nodes}; the itinerary planner, for example, contributes its planners, a ranker and a selector. Each node exposes a contract $C_n=(I_n,O_n,\mathrm{Pre}_n,\mathrm{Post}_n,\widehat{\mathrm{Claims}}_n,\mathrm{Propose}_n,\mathrm{Score}_n)$: typed inputs and outputs, preconditions, guaranteed postconditions, an ex-ante claim model, a proposal operation and an optional preference score. The claim model gives, before the node runs, the distribution $\widehat c_{nP}(\cdot\mid s)$ of its claim on pool $P$ in typed state $s$. $\mathrm{Propose}_n(s)$ may run the capability's full search but changes no shared state; it returns a candidate $x_n$, evidence for its postconditions, and the candidate-conditioned claim $\widehat c_{nP}(\cdot\mid x_n,s)$. An edge from $i$ to $j$ is executable only when $\mathrm{Post}_i$ implies $\mathrm{Pre}_j$, possibly through a reconciliation node. Scoring nodes rank feasible alternatives but cannot admit infeasible ones, and feasibility nodes admit or reject without encoding preference.

Edges are typed transformations. A \emph{projection} maps many values to one, such as a city's attractions to a price tier; a \emph{derivation} computes a fact from committed decisions, such as hotel nights and daily time windows from travel times. Derived facts feed several components, so they belong to the composition: an arrival at 10:30 and a departure four days later at 18:00 imply four hotel nights but five days of activities. Each unit of consumption has exactly one owning claim, so derivations and projections can condition a claim but never duplicate it. Resolution works backward from the requested outputs, finds a provider for every input and builds an acyclic dependency graph $G_0$; pools and allocation-feedback edges are then attached (lines 1--2). A request for all three capabilities has 13 nodes: three transformations (city, time windows, price tier), the three fast itinerary planners (greedy, A*, beam) with a ranker and a selector, the budget planner, a mapping from attractions to packing activities, and three packing nodes. The first nine form the money closure of the Lisbon request.

\begin{algorithm}[t]
\caption{Gated execution of a resolved request}
\label{alg:resolve}
\footnotesize
\begin{algorithmic}[1]
\Require typed goal $g$; capability library $L$; pool capacities $B_P$
\State $G_0\gets\textsc{ResolveTypedDAG}(g,L)$; if a semantic cycle cannot be composed, \Return clarification or no plan
\State $G\gets\textsc{AttachPools}(G_0,\widehat{\mathrm{Claims}},B)$; $A\gets\emptyset$
\While{$U=\textsc{Unresolved}(G,A)\neq\emptyset$}
\State compute the margins $m_P(s)$ of all pools over $A$ and $U$
\If{$\mathcal B=\{P:m_P(s)<0$ or $P$ fails certification$\}\neq\emptyset$}
\State $H\gets\textsc{ResourceClosure}(U,\mathcal B)$; $J\gets\textsc{RegisteredJoint}(H)$
\State \textbf{if} no $J$ \textbf{then} \Return clarification or no plan
\State solve $J$ to its deadline; re-gate every pool under the solution
\State \textbf{if} no result certifies $\bigwedge_{n\in H}\mathrm{Post}_n$, $m_P\!\ge\!0$ and certification \textbf{then} \Return no plan
\State accept $H$ provisionally, keeping conditioned claims; $A\gets A\cup H$
\Else
\State $(x_n,e_n,c_n)\gets\textsc{Propose}(\text{first executable }n\text{ in topological order of }G_0[U])$
\State substitute $\widehat c_{nP}(\cdot\mid x_n,s)$; re-gate every pool; $\mathcal B'\gets$ pools newly binding or failing certification
\If{$\mathcal B'=\emptyset$ and $e_n\models\mathrm{Post}_n$}
\State accept $n$ provisionally, keeping its conditioned claim; $A\gets A\cup\{n\}$
\Else
\State discard the proposal; \textbf{if} $e_n\not\models\mathrm{Post}_n$ \textbf{then} \Return contract violation
\State solve the closure of $\mathcal B'$ as in lines 6--10
\EndIf
\EndIf
\EndWhile
\State \textbf{if} $\neg\textsc{ValidateAllContracts}(G,A)$ \textbf{then} \Return no plan
\State \textbf{if} some uncertain pool fails certification on fresh scenarios \textbf{then return} no plan
\State \Return the composed plan
\end{algorithmic}
\end{algorithm}

\subsection{Slack Gate}
\label{sec:slack}
Treating every feedback cycle as a reason to solve jointly would solve jointly almost always. The slack gate instead asks, at each step, whether a shared resource is tight enough that decomposition could break its limit. Acceptance fixes a decision, not a realization, so accepted money claims stay random; only consumption that is deterministic once the context is fixed (time and space) enters a ledger $d_P^{\mathrm{det}}(s)$. With $\mathcal{A}^{\mathrm{st}}(s)$ the accepted stochastic claimants and $U(s)$ the unresolved nodes,
\begin{gather}
\widehat C_P^{\mathrm{tot}}(s)=d_P^{\mathrm{det}}(s)+\!\!\sum_{n\in\mathcal{A}^{\mathrm{st}}(s)}\!\!\widehat c_{nP}(\cdot\mid x_n,s)+\!\sum_{n\in U(s)}\!\widehat c_{nP}(\cdot\mid s),\notag\\
m_P(s)=B_P-Q_\alpha\big(\widehat C_P^{\mathrm{tot}}(s)\big).
\label{eq:margin}
\end{gather}
Pool $P$ binds when $m_P(s)<0$, and an uncertain pool also binds when its total fails certification (§\ref{sec:guarantee}). Every claim appears once in the total, the quantile is taken of the joint total rather than summed from per-node quantiles, and the shared demand factor linking accepted and unresolved claims survives acceptance. Collapsing accepted claims to point values would shrink the modeled variance with every acceptance while the true variance persisted, leaving the gate most permissive exactly when the pool is tightest. For Lisbon, the total adds the budget plan's flights and hotel to the itinerary's attractions, all driven by one demand factor, and money binds when the 90\% quantile exceeds \$2,800. A non-negative margin gives the pool's chance constraint only if the quantile is exact; the gate estimates it from a fixed sample of scenarios, so the statement holds up to sampling error, which certification addresses (§\ref{sec:guarantee}). The guarantee is per pool; a union bound gives joint coverage of at least $1-|\mathcal P|(1-\alpha)$.

Binding is a prediction from estimates, not a proof that coordination is necessary: a negative margin can open a closure that sequential execution would have survived, and a non-negative margin does not ensure that every later node still fits without reopening an accepted one.

\noindent\textbf{Gate, Then Propose.} The gate runs before each proposal (line 4), so no node runs its full search when the ex-ante total already binds, and ex-ante claims must be cheap to obtain. A clear margin licenses only the next proposal: its conditioned claim replaces the ex-ante claim in the joint model and every pool's margin is recomputed (line 13); a proposal that makes a pool bind is discarded and the closure is solved instead. Conservative ex-ante claims can cause needless binds, which we measure (§\ref{sec:rq2}).

\subsection{Resource Closures}
\label{sec:closures}
When pools $\mathcal B$ bind, \textsc{ResourceClosure} collects their unresolved claimants and the unresolved nodes these depend on, and nothing else. For Lisbon, the closure holds the itinerary planners, ranker and selector, the budget planner and the transformations they read. Packing claims only space and depends on the itinerary rather than the reverse, so it runs afterward and is re-gated on space. A carry-on ski trip binds only space; its closure lies inside the packing capability, whose CP-SAT contract already covers it. If every unresolved node claims the binding pool, the system falls back to always-joint solving. A registered composite solver must cover the closure's decisions and return a solution satisfying every constituent postcondition; the solution is re-gated (line 8), since the cross-node constraint belongs to no local contract. Without a registered solver, or without a passing incumbent by the deadline, the system fails closed with ``no plan''.

The gate is also a personalization decision. When no pool binds, each capability optimizes its own preference in turn, and levers that trade one capability against another, such as planning below the city's price tier, are not tried. When a pool binds, the composite solver reconciles the capabilities' preferences over the closure's joint decisions.

\subsection{Certification and Feasibility Guarantee}
\label{sec:guarantee}
The gate estimates each quantile from a fixed sample of cost scenarios, and a plan chosen as the best on a sample looks safer on that sample than it is (§\ref{sec:rq1}). The orchestrator therefore tests every uncertain pool on a second sample of $n$ scenarios drawn from $M$ independently of the gate's, and a plan passes when the one-sided Clopper--Pearson upper bound on its overrun probability~\cite{clopper1934}, at confidence $1-\delta$, is at most $1-\alpha$. The test decides acceptance at every step: a pool binds when its total fails (line 5), a proposal is accepted only if it passes (line 13), the budget node takes the most convenient frontier plan that passes, and a composite solver whose setting fails moves to cheaper settings until one passes (line 9). A failure therefore leads to a cheaper certified plan rather than to ``no plan'', and the final test (line 23) releases nothing that has not passed.

Feasibility rests on five assumptions. (A1) Each capability satisfies its declared postconditions. (A2) Every reconciliation is valid. (A3) The gate's quantile for each stochastic pool is the exact $\alpha$-quantile under the joint cost model. (A4) Every sequential proposal is re-gated before acceptance. (A5) A composite solver returns, by its deadline, a solution verified on the constituent postconditions, every pool's margin and certification, or nothing.

\textsc{Proposition 1.} \emph{(i) Under A1--A5, every provisionally accepted prefix satisfies the declared deterministic postconditions, and every stochastic pool satisfies its level-$\alpha$ chance constraint under the cost model. (ii) Without A3, a plan whose overrun probability under the cost model exceeds $1-\alpha$, fixed before the certification scenarios are drawn, passes certification with probability at most $\delta$. No plan is released unless final contract validation and certification succeed.}

\emph{Proof sketch.} Induct over acceptances: sequential steps validate postconditions and re-gate every pool; composite steps verify all constituent postconditions and re-gate under the joint decision; under A1--A5 these preserve feasibility. For a fixed plan, certification overruns are binomial, and Clopper--Pearson coverage~\cite{clopper1934} bounds erroneous acceptance by $\delta$.

A3 does not hold exactly, because the gate selects plans on a sampled quantile, so certification, not the gate, carries the statistical guarantee. Part (ii) covers a plan that passes its first test. A plan chosen after a failure is itself selected on the certification scenarios, so the bound holds for each candidate tested, not for the one released. This is common: certification changes the released plan for 907 of 1,584 releases (57\%); each of the 352 plans the validator rejects without certification is either replaced by a certified plan (351) or withheld (1). The validator rejects none of these changed plans (highest overrun probability 9.4\%), because the test passes a plan only if it overruns in at most 8.85\% of the certification scenarios; in the synthetic benchmark, where many certified candidates compete, the validator does reject some, and a small margin at certification absorbs them (§\ref{sec:rq2}). Neither part covers a cost model whose prices are wrong, and the proposition guarantees neither completeness nor optimality: closures optimize unresolved decisions given accepted ones and never backtrack, so a different early choice could admit a plan where the system answers ``no plan''. Section~\ref{sec:rq3} measures this loss.

\subsection{Instantiation for Travel}
\label{sec:instantiation}
The system runs end to end against the three capabilities' own code. The money pool's cost model is the budget planner's own~\cite{dev2026budgeteer}: each cost line has a lognormal price fitted to its 10th, 50th and 90th percentiles, adjusted for booking lead time and the planner's dependency rules, and one shared demand factor couples the lines, with loadings from 0.1 (documents) to 0.7 (flights). The percentiles are synthetic baselines for 99 U.S. destinations, audited against public price sources, and the itinerary's attraction spend is folded into the same model. The gate computes margins on a fixed sample of 2,000 scenarios at $\alpha=0.9$, and certification draws $n=2{,}000$ further scenarios. The itinerary's ex-ante money claim comes from its fastest planner at the current price tier, and the budget's from its cheapest frontier plan.

Two composite solvers are registered. The packing composite is the packing system's CP-SAT model. The itinerary and budget composite, written for this composition, searches three levers: planning one or two price tiers below the city's, choosing a lower-ranked itinerary, and moving along the budget frontier, which it binary-searches since cost falls monotonically along it. Among the settings that pass certification, it returns the one maximizing $U=0.5\,\tilde r+0.5\,v$, where $\tilde r$ is the itinerary reward min-max normalized over the feasible settings and $v\in[0,1]$ the budget plan's convenience. Planners are seeded per call, so the same request always gives the same plan.

\subsection{Relaxations, Replanning and Language}
\label{sec:recovery}
\label{sec:language}
A ``no plan'' answer is useful only with a way forward. The gate already knows which pool bound and by how much, so the system searches, for each dimension on its own, the smallest change that gives a plan: the budget is bisected to within 0.2\%, days are removed from the end, the month moves up to six months either way, and a bag limit can grow to the next larger bag. Each candidate is checked by running the whole system, so every offered relaxation comes with a plan.

Released plans are stored with every accepted decision. A replan applies a change, such as a budget cut, a day fewer, one more person, rain on one day or a closed attraction, and re-runs the same gate and solvers with a warm start: a node whose inputs and levers are unchanged returns its previous decision. A budget cut that still fits keeps the whole plan, and one that makes money bind re-opens exactly the money closure, so closures bound replanning as they bound solving. A closed attraction is replaced in its time slot by the most popular open attraction that fits. Optionally, a re-opened joint solve prefers stability, returning among certified settings within $\varepsilon$ of the best utility the one most like the previous plan.

The parser turns a request into a typed goal, with a pattern parser, a language model, or a hybrid in which the model extracts fields, every value must be traced to the user's words, and the pattern parser cross-checks (§\ref{sec:rq5}). Unresolved numeric, temporal or destination ambiguity triggers a clarification question rather than a plan for the wrong contract. Follow-ups are parsed against the earlier plan: ``rained out on day 2 and 10\% over budget'' becomes a rain exclusion on day 2 and a 10\% budget cut, run as a warm replan. Explanations are rendered from the executor's own values without running a planner, and a language model may reword an answer only if the rewording carries exactly the same numbers.

\begin{figure*}[t]
\begin{minipage}[t]{0.32\textwidth}
\centering
\includegraphics[width=\linewidth]{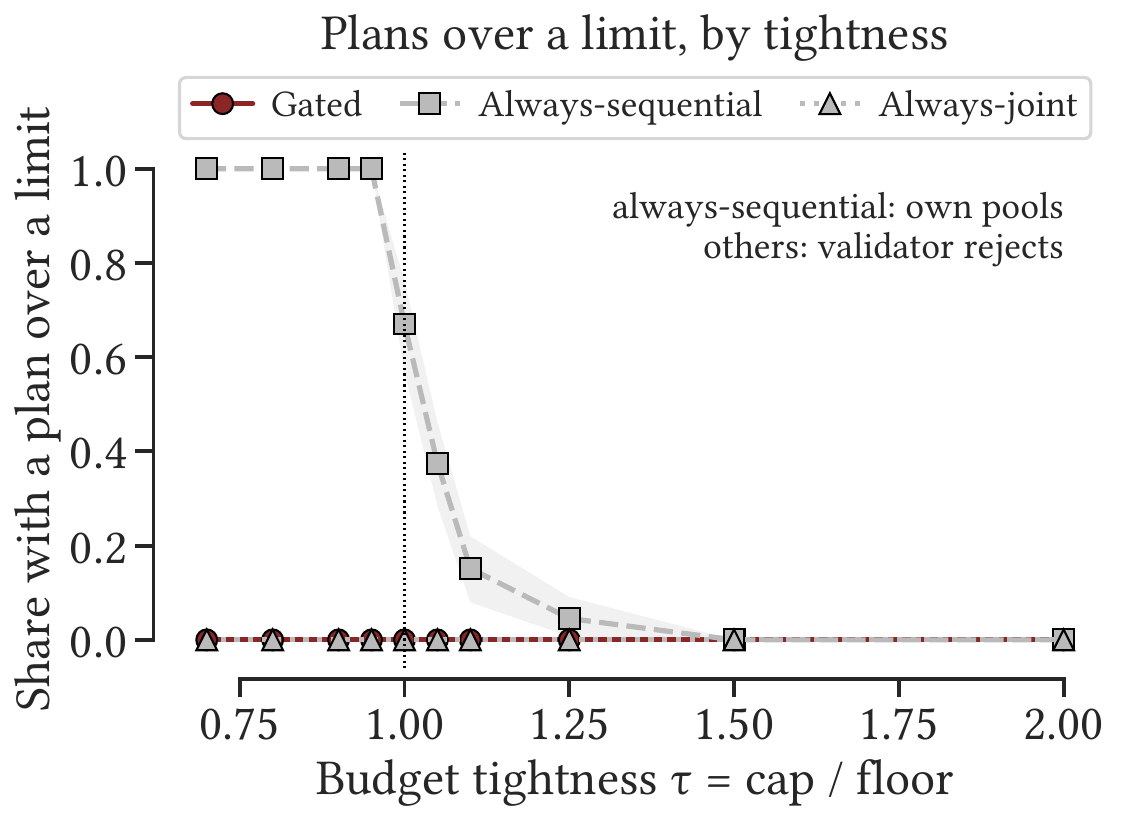}
\caption{Always-sequential plans break a limit for most requests up to $\tau=1.05$; no certified plan does.}
\label{fig:overlimit}
\end{minipage}\hfill
\begin{minipage}[t]{0.32\textwidth}
\centering
\includegraphics[width=\linewidth]{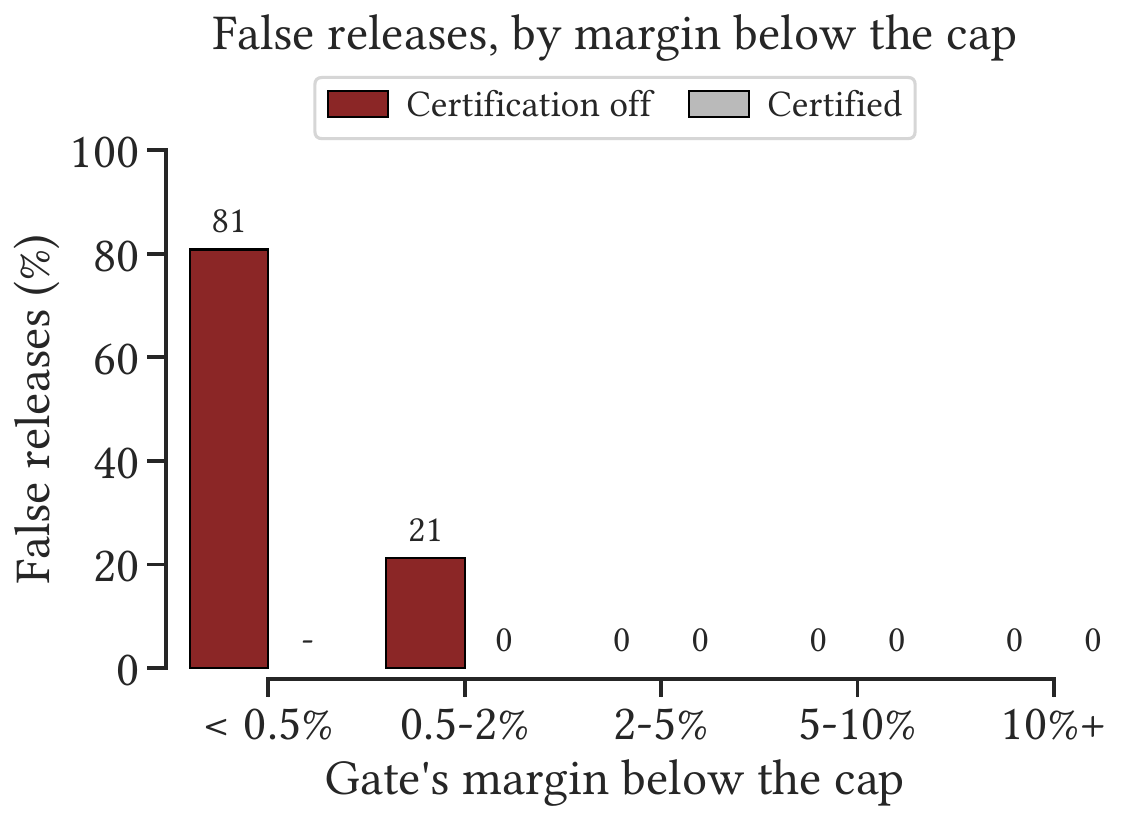}
\caption{With certification off, plans within 2\% of the cap often fail (81\% within 0.5\%); certified plans never fail.}
\label{fig:wincurse}
\end{minipage}\hfill
\begin{minipage}[t]{0.32\textwidth}
\centering
\includegraphics[width=\linewidth]{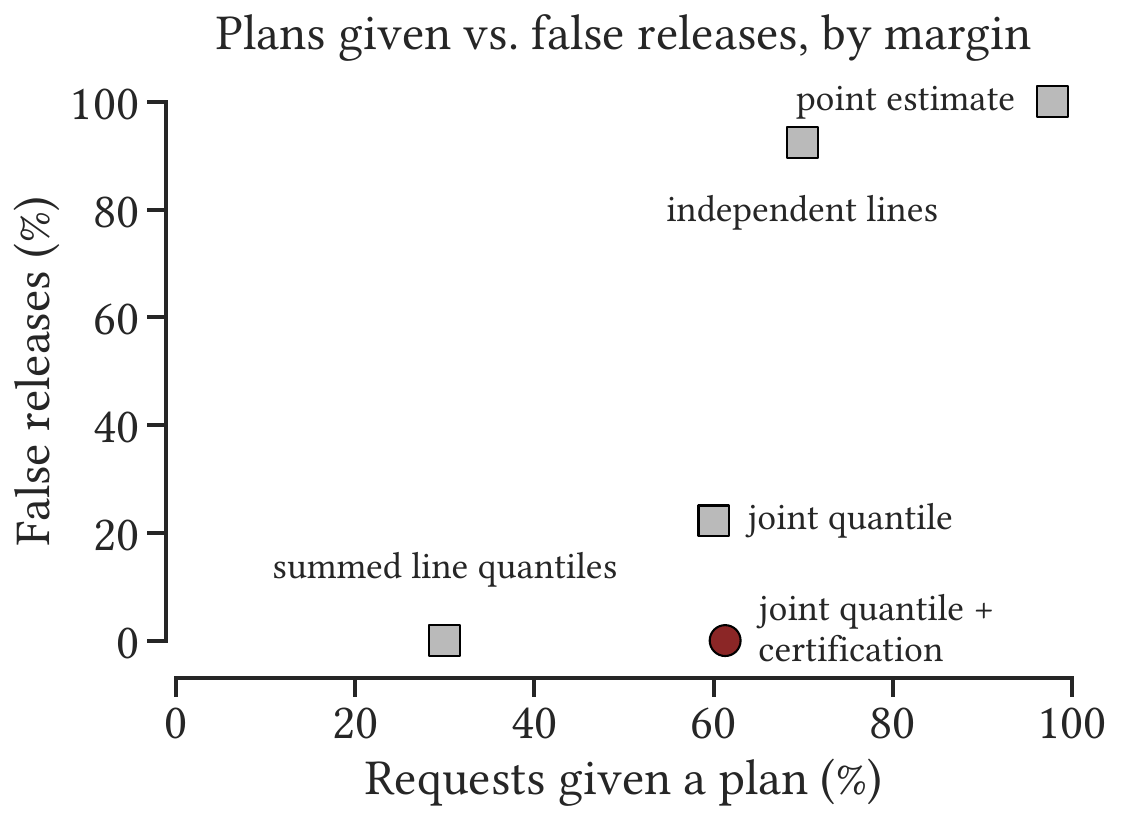}
\caption{Per-component margins (certification off) overrun or lose plans; a certified joint quantile does neither.}
\label{fig:margins}
\end{minipage}
\end{figure*}

\section{Evaluation Setup}
\label{sec:setup}
We describe the requests, the methods we compare and how plans are judged, then the studies of timing, scale, recovery and language.

\noindent\textbf{Requests.} We sample one trip (origin, destination, month, length and party size) in each of the 88 destination cities of the budget system's catalog, and ask each as four request types written from one template (Table~\ref{tab:taxonomy}): \emph{atomic} (an itinerary only), \emph{joint budget} (itinerary and budget plan under a cap), \emph{allocation} (all three under a cap) and \emph{conflict} (all three under a cap and a bag limit). Budgets are set by \emph{budget tightness} $\tau$, the cap divided by the trip's \emph{feasibility floor}, the smallest cap at which the always-joint system releases a certified plan, found by bisection to 0.2\%. We use $\tau\in\{0.7,0.8,0.9,0.95,1,1.05,1.1,1.25,1.5,2\}$, so a certified plan exists at every level from $\tau=1$ upward by construction. The grid has no level between $\tau=1$ and $1.05$, where certified plans have little slack, so the evaluation asks whether the other methods match always-joint on this grid, and at what cost. This gives 31 requests per trip, 2{,}728 in total, 2{,}640 with a budget. Because the grid places most budgets near the floor, we also weigh results by the 83 budgets participants chose in the budget system's user study (median $\tau=1.51$).

\noindent\textbf{Request Types and Coupling.} The types invite different couplings (Table~\ref{tab:taxonomy}). An atomic request suggests none, while a cap shared by several components (joint budget, allocation) or combined with a bag limit (conflict) suggests pools that may bind. Follow-ups about an earlier plan form three more types: a \emph{what-if} edit, a \emph{replanning} request after something changed, and an \emph{explanation}. The table shows why the type serves only as a prior (§\ref{sec:taxonomy}): within every budgeted type, binding follows how tight the cap is for that trip. All three bind on every request at or below the floor and on 82\% to 92\% at 1.1 times it, while at 1.5 times joint-budget requests never bind and allocation requests almost never do; conflict requests still bind in 59\% of cases, through the bag rather than the budget.

\noindent\textbf{Methods.} \emph{Gated} is the system of §\ref{sec:arch}. \emph{Always-sequential} runs the same nodes in topological order with no gate, each capability optimizing alone. \emph{Always-joint} solves every request as one closure over all its nodes, with the composite solver's levers and objective. \emph{Validate-and-repair}, the reactive pattern of LLM-Modulo systems~\cite{kambhampati2024llmmodulo}, runs the nodes sequentially, checks the finished plan against every pool and the certification, and only when a limit breaks solves the binding pool's closure. \emph{Static splits} reserve a share $s\in\{5,10,20,30\%\}$ of the cap for attractions and the rest for the budget plan, each keeping its share at its own 90\% level. All methods use the same three fast planners. The \emph{exhaustive oracle} enumerates every combination of price tier, ranked itinerary and frontier budget plan per trip and request type, and judges feasibility with the gate's cost model and certification threshold; it shares the composite solver's search space, so it tests whether selective solving and early commitment lose plans or utility within that space. Per trip and request type it covers price-tier steps 0--2, the ranked itinerary candidates and up to 256 frontier budget plans, recording each combination's money quantile and certification threshold once, independent of the cap; the median space holds 450 combinations (at most 675), enumerated in 22.1 s over the same three planners for every trip and 53.0 s over all five for 20 of the 88 trips.

\noindent\textbf{Validation and Robustness.} An independent validator re-simulates every released plan's money with a different seed and 100,000 scenarios, failing it if the overrun probability exceeds 10\% (a 95\% Wilson interval~\cite{wilson1927} marks clearly over or within), and recounts days and bag weights from the plan. Replaying the gate's own settings reproduces its 90\% cost exactly on all 17{,}952 validated plans, so differences come from the sample, not from a reconstruction error. The validator shares the cost model, so it measures sampling error only. A stress test re-asks the 2{,}640 budgeted requests with one setting changed: with certification off, the money-estimate bias, $\alpha$, the gate's sample size and the way the margin is computed; anything chosen from these runs is chosen on 44 calibration cities and reported on the other 44. Robustness re-judges released plans under costs the model does not assume: the spreads of the budget system's synthetic trip-cost records, spreads 25\% and 50\% wider, and prices that move together more strongly.

\noindent\textbf{Timing.} Times come from a separate study on an idle 4-core machine (310 requests of 10 trips, every method, three shuffled repetitions with cold caches); the core runs shared one machine, so we use them for proposal counts only.

\noindent\textbf{Synthetic Benchmark.} A synthetic benchmark reuses the orchestrator's contracts, executor, gate and certification on random systems of $k$ capabilities with $L$ options each. Each capability claims a random subset of up to five pools, with at least two claimants per pool; dependencies range from none to chains, trees and DAGs; and uncertain costs are lognormal with a shared shock per pool. The base system has $k=10$, three pools, $L=16$, DAG dependencies and money as its only uncertain pool. Varying $k$ from 3 to 20, the pool count, $L$, the topology and which pools are uncertain gives 19 configurations and 950 instances (ten systems and five caps each). The gate uses 200 scenarios, certification 2,000 and validation 20,000; CP-SAT has a 60 s deadline.

\noindent\textbf{Recovery and Language.} Relaxations are evaluated on the 1{,}056 budgeted requests the gated system answered ``no plan'', and replanning on 14{,}960 replans of 616 released plans, each perturbed in up to eight ways (budget cuts of 5, 10, 20 and 30\%, a day fewer, one more person, rain on one day and a closed attraction) and replanned warm, warm with the stability preference ($\varepsilon=0.05$), and from scratch. Parsing uses 300 constructed requests with gold fields, and follow-ups 17{,}072 templated messages.

\noindent\textbf{Statistics.} Requests of one trip are not independent, so intervals in brackets are 95\% bootstrap intervals over the 88 destination cities, each one trip with its 31 requests (2,000 replicates).

\noindent\textbf{Reproducibility.} Experiments write raw records, analysis computes every statistic, and every number and figure in the paper is generated from sealed, versioned records. Planners, cost scenarios and CP-SAT use fixed seeds, so the same request always gives the same plan.

\section{Results}
\label{sec:results}
Each research question asks a decision a builder of such a system faces, and answers it against the alternative the builder would otherwise choose.

\subsection{RQ1: Feasibility}
\label{sec:rq1}
RQ1 asks what it takes to keep composed plans within their shared limits. We measure the share of plans over a limit, by their own pools and by the validator, then switch certification off to see what it removes, and change the cost model to see what it cannot.

\begin{figure*}[t]
\begin{minipage}[t]{0.32\textwidth}
\centering
\includegraphics[width=\linewidth]{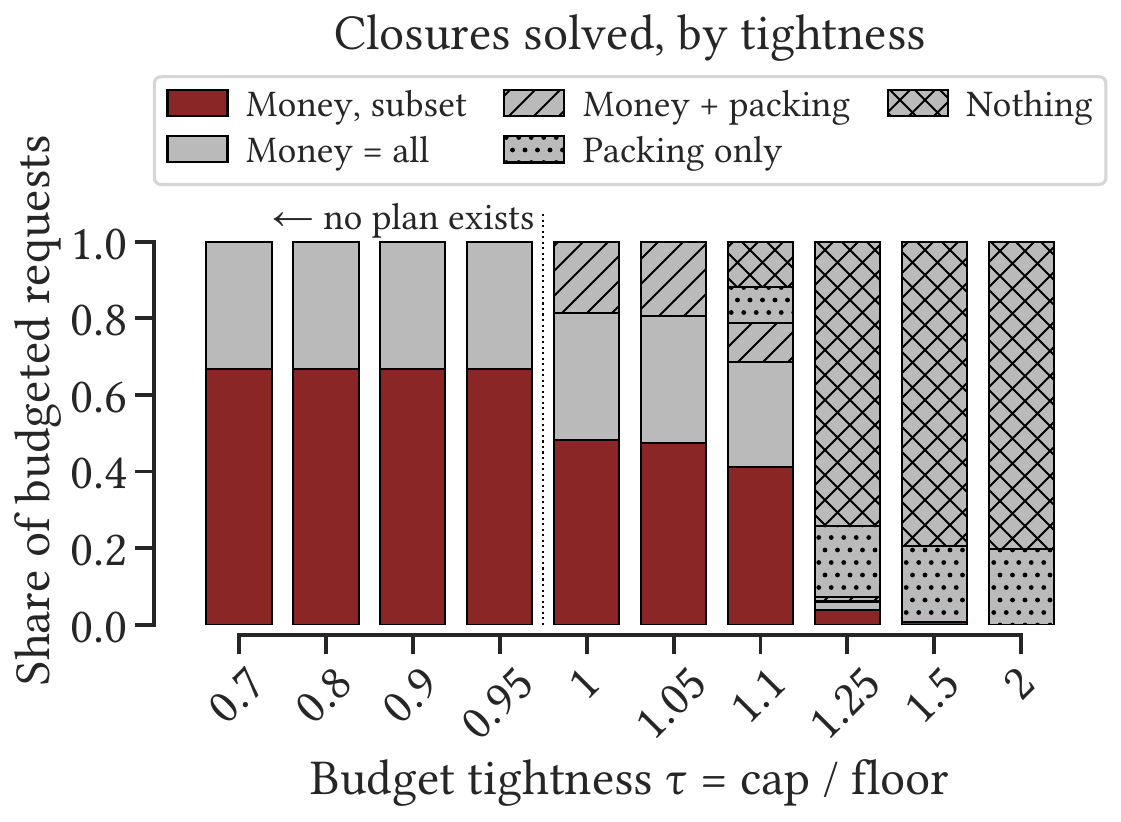}
\caption{Near the floor the money closure is solved; above it nothing is, except packing closures in conflict requests.}
\label{fig:closuremix}
\end{minipage}\hfill
\begin{minipage}[t]{0.32\textwidth}
\centering
\includegraphics[width=\linewidth]{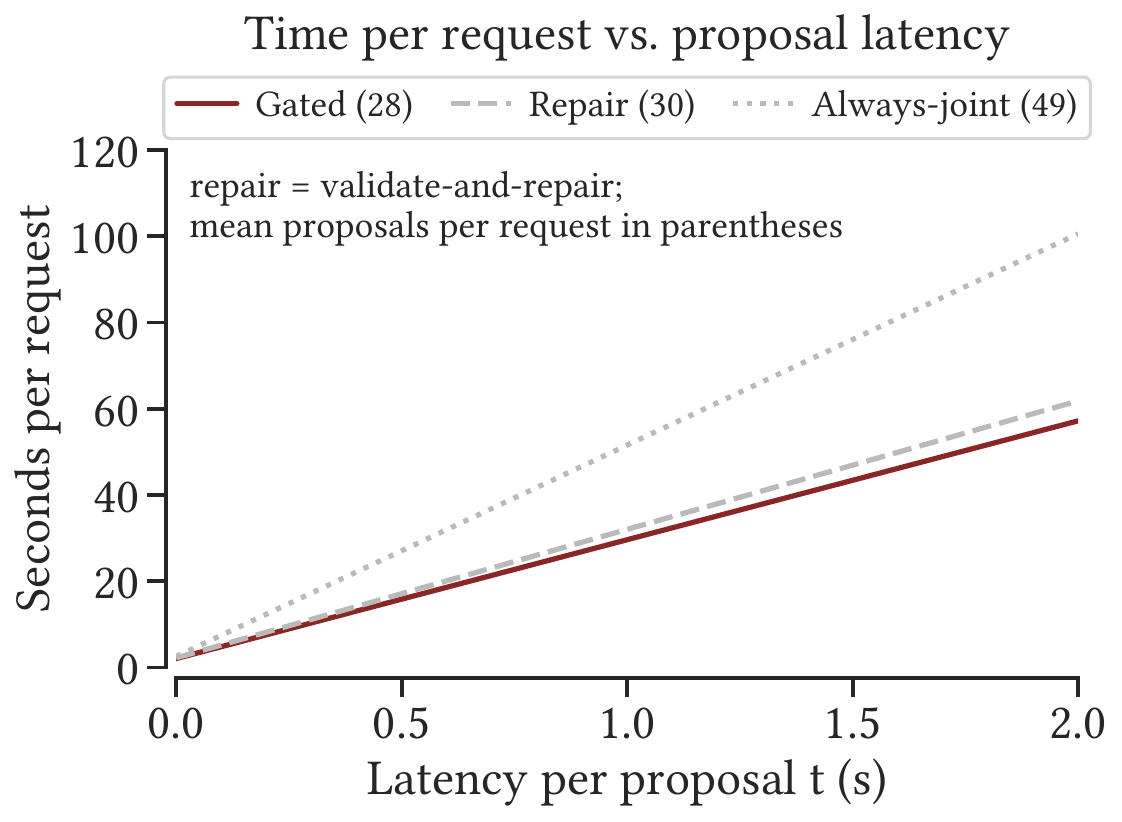}
\caption{With a latency $t$ per proposal, gating is the fastest at every $t$; always-joint takes 1.7 times as long.}
\label{fig:costlatency}
\end{minipage}\hfill
\begin{minipage}[t]{0.32\textwidth}
\centering
\includegraphics[width=\linewidth]{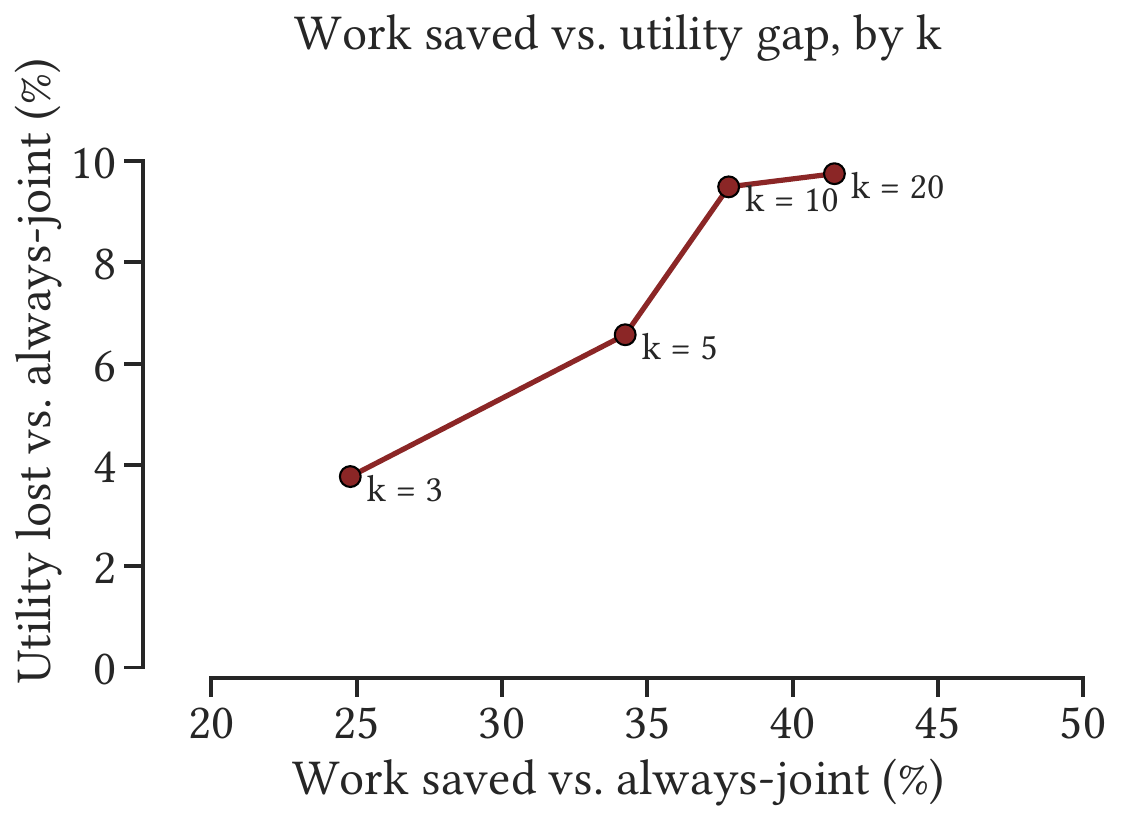}
\caption{In synthetic systems of 3 to 20 capabilities, gating saves more work but loses more utility.}
\label{fig:scaling}
\end{minipage}
\end{figure*}

\noindent\textbf{Running the Capabilities One After Another Breaks Limits.} Always-sequential execution releases a plan for every request, and 51\% of them break a shared limit. Below the floor every one does; at or above it 20.7\% [16.8--24.7\%] still do, 67\% at the floor and 4.5\% at $\tau=1.25$ (Figure~\ref{fig:overlimit}). The overshoots are not marginal: among the 1{,}584 requests at or above the floor, 328 plans exceed the cap at the 90\% level, by a median of 5\% and up to 47\%. Weighted by the budgets users chose, most of them well above the floor, 5.2\% of budgeted requests still break a limit. Dividing the cap in advance does not fix this. Static splits put a plan over a limit for 40.4\% to 52.9\% of budgeted requests, and between $\tau=1$ and $1.1$ a 5\% share for attractions is over in 1\% of requests and a 30\% share in 43\%, so a fixed split works only at the share it was tuned to.

\noindent\textbf{Certified Releases Stay Within Every Limit.} The gated system releases 1{,}672 plans and answers ``no plan'' to 1{,}056 requests, all below the floor, where the oracle agrees (§\ref{sec:rq3}). The validator rejects none of its releases, and none of always-joint's or validate-and-repair's (Figure~\ref{fig:overlimit}). It finds the gate's model slightly optimistic, with a higher overrun probability for 93\% of gated releases by a median of 0.54 points; certification absorbs that difference, and the highest overrun probability the validator finds is 9.4\%. A gated release's \emph{expected overrun}, the mean amount by which its cost exceeds the cap across the validator's scenarios, is at most 1.6\% of its cap (\$144); for always-sequential plans over the limit it is 4.6\% (\$105) at the median and up to 54\%.

\noindent\textbf{Without Certification, One Release in Five Fails.} With certification switched off and everything else equal, the validator rejects 22\% [20--24\%] of the gated system's releases. These \emph{false releases} are plans chosen as the best on the gate's own sample, and therefore lucky on it: 81\% of the releases within 0.5\% of the cap fail, 21\% of those 0.5--2\% inside it, and 0\% of those further inside (Figure~\ref{fig:wincurse}). The more candidates a release was chosen from, the more often it fails (18\% with none, 30\% with 10 to 50). Certifying each plan on scenarios that played no part in choosing it removes these failures without losing a valid plan: the share of requests given a plan stays at 60\%.

\noindent\textbf{The Margin Must Be Computed Jointly.} Per-component safeguards fail (Figure~\ref{fig:margins}). Judging money by point estimates releases plans the validator rejects in 100\% of cases, and treating the cost lines as independent in 92\%. Summing per-line 90\% quantiles never overruns but gives a plan to only 30\% of requests, against 60\% for one quantile of the joint total.

\noindent\textbf{A Small Margin Also Works; the Confidence Level Does Not.} Without certification, a margin $b=2\%$ on the money estimate, chosen on the calibration cities, leaves 0\% false releases on the test cities at no loss of plans. Raising $\alpha$ is no substitute: the false-release rate is 5\%, 12\%, 22\%, 25\%, 12\% for $\alpha$ from 0.8 to 0.99, low only where few requests get a plan (Table~\ref{tab:bias}). A stricter $\alpha$ moves the boundary, but releases still concentrate just inside it.

\begin{table}[hbt]
\centering\footnotesize
\setlength{\tabcolsep}{3pt}
\caption{With certification off: false releases by money-estimate bias $b$ and by $\alpha$, 2{,}640 requests per setting.}
\label{tab:bias}
\begin{tabular}{@{}llcccc@{}}
\toprule
Setting & Value & \thead{given\\a plan} & \thead{false\\releases} & \thead{false-release\\rate} & \thead{needless\\binds} \\
\midrule
bias $b$ & -5\% & 0.70 & 1776 & 96\% & 13.6\% \\
 & +0\% & 0.60 & 352 & 22\% & 18.6\% \\
 & +1\% & 0.60 & 27 & 2\% & 20.2\% \\
 & +2\% & 0.60 & 0 & 0\% & 21.2\% \\
 & +5\% & 0.50 & 0 & 0\% & 23.1\% \\
 & +10\% & 0.40 & 0 & 0\% & 24.6\% \\
\midrule
level $\alpha$ & 0.8 & 0.80 & 106 & 5\% & 11.2\% \\
 & 0.85 & 0.70 & 225 & 12\% & 13.6\% \\
 & 0.9 & 0.60 & 352 & 22\% & 18.6\% \\
 & 0.95 & 0.41 & 274 & 25\% & 24.0\% \\
 & 0.99 & 0.23 & 73 & 12\% & 32.4\% \\
\bottomrule
\end{tabular}
\end{table}

\noindent\textbf{Certification Fixes Sampling Error, Not Model Error.} Certification checks a plan against the same cost model on new draws. Under the spreads of the budget system's synthetic trip-cost records, every gated release stays within its limit (100\%). If costs spread 25\% wider than the model believes, only 0.7\% do, and 18\% if prices move together more strongly; always-joint and validate-and-repair fare the same. Static splits look more robust (60\% for a 30\% split under the wider spreads) only because they leave part of the cap unspent.

\takeaway{Running the capabilities one after another breaks a shared limit for 51\% of requests. The gated system, always-joint and validate-and-repair release no plan an independent re-simulation rejects, because of release certification: without it 22\% of releases fail. The margin must be computed on the joint total, and the remaining risk lies in the cost model.}

\subsection{RQ2: Efficiency}
\label{sec:rq2}
RQ2 asks when coordinating pays. We report how often pools bind by tightness and how large their closures are, the time per request on an idle machine and with a latency $t$ per proposal, and the effect of more capabilities.

\begin{figure*}[t]
\begin{minipage}[t]{0.32\textwidth}
\centering
\includegraphics[width=\linewidth]{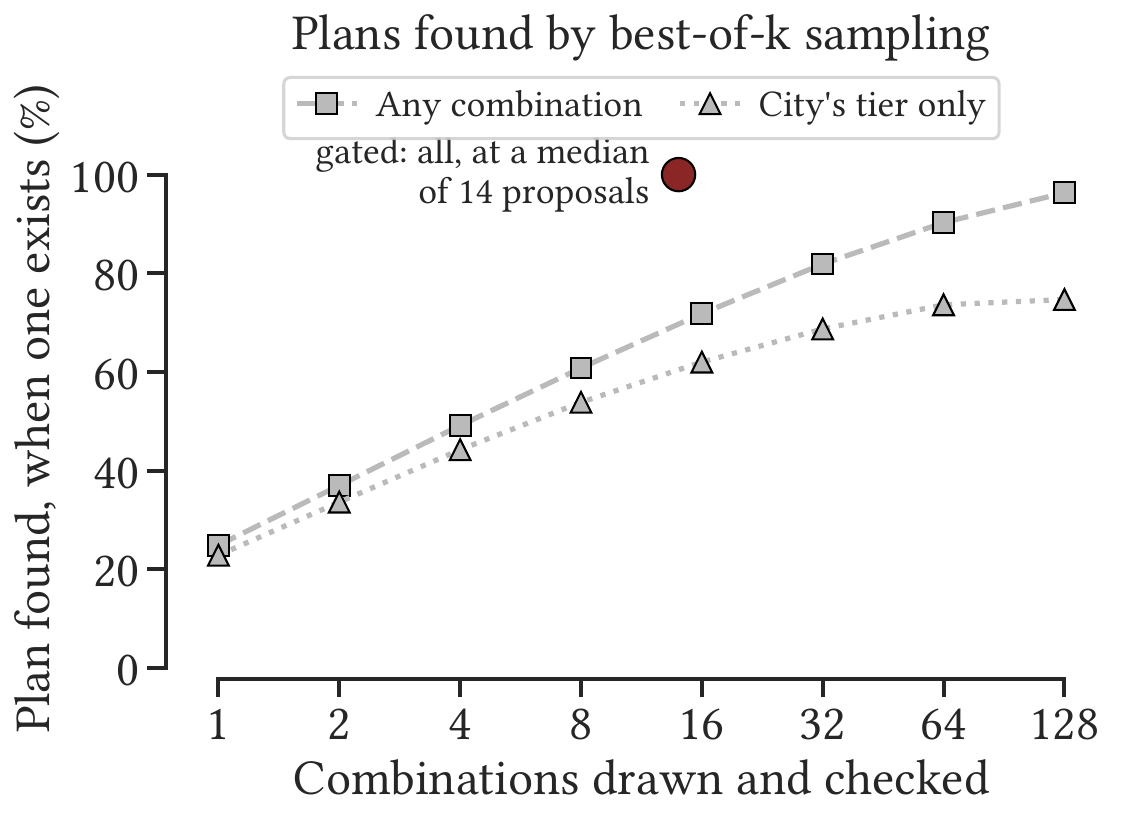}
\caption{Best-of-$k$ sampling needs 128 draws to find 96\% of plans; gating finds all with a median of 14 proposals.}
\label{fig:bestofk}
\end{minipage}\hfill
\begin{minipage}[t]{0.32\textwidth}
\centering
\includegraphics[width=\linewidth]{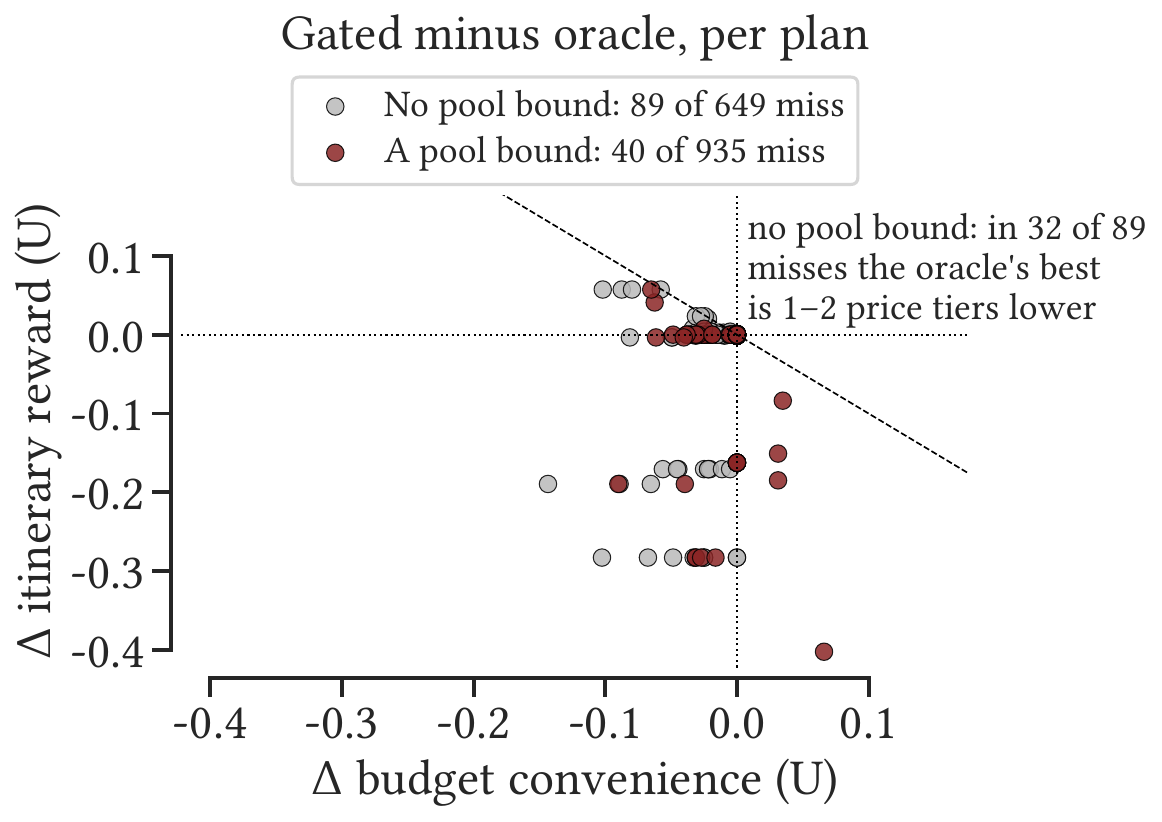}
\caption{Gated plans that miss the oracle's best mostly have a worse itinerary, from a higher price tier.}
\label{fig:components}
\end{minipage}\hfill
\begin{minipage}[t]{0.32\textwidth}
\centering
\includegraphics[width=\linewidth]{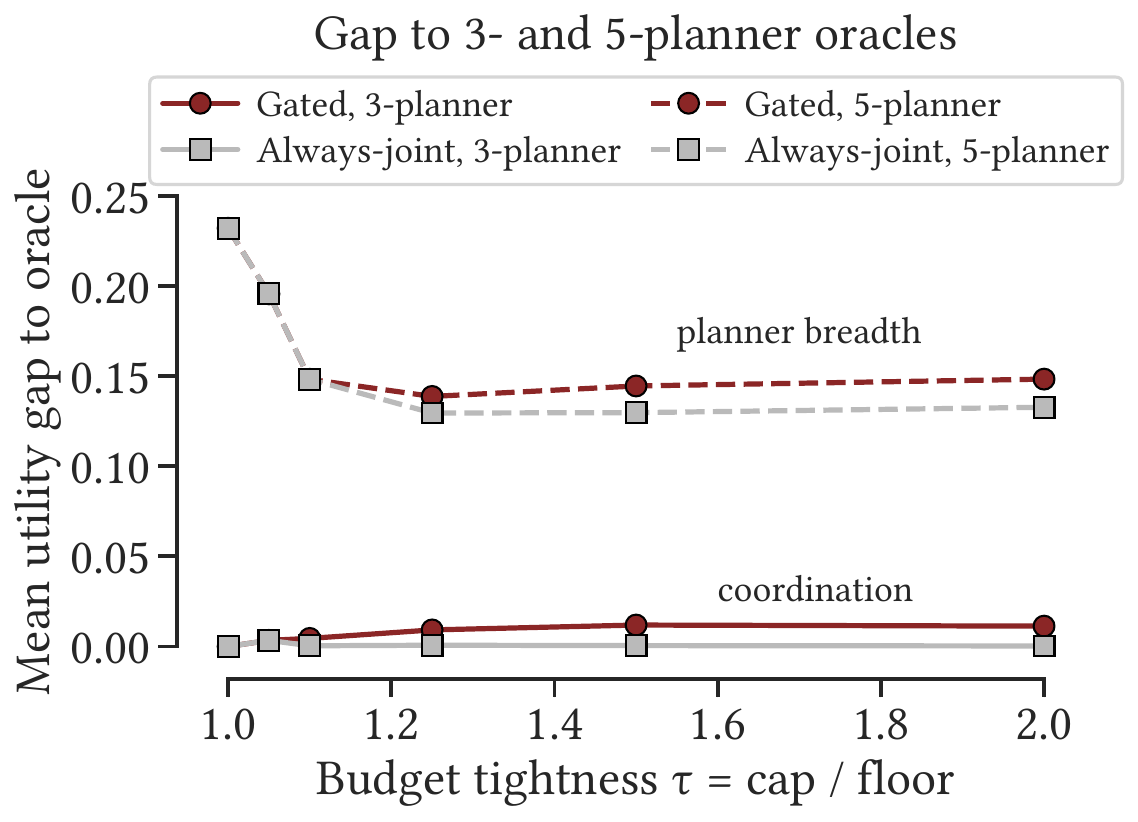}
\caption{At every tightness, missing planners cost ten times more utility than selective coordination.}
\label{fig:decomp}
\end{minipage}
\end{figure*}

\noindent\textbf{Binding Follows Budget Tightness.} Some pool binds for every budgeted request at $\tau\le1.05$, for 26\% at $1.25$ and for 20\% at $1.5$ and above (Figure~\ref{fig:closuremix}). The binds far above the floor are space binds in conflict requests, the only type with a bag limit, so request type matters only through what a type claims. Follow-ups show the same: preference-only edits, which a structural prior would run sequentially, open the money closure in 39\% of cases, against 46\% for edits that change money or time claims, and what-ifs open it in 70\% of cases at $\tau=1.1$ but 22\% at $1.5$.

\noindent\textbf{Closures Are Proper Subsets.} Over 2{,}122 binding events the median closure covers 0.69 of the unresolved graph: the money closure of 9 nodes, with packing's four nodes run afterward.

\noindent\textbf{No Missed Limits, Some Needless Binds.} On the 737 requests where the gate never bound, 0 released plans break a limit, so the gate misses no limit; on 6.4\% of them always-joint finds a plan better on both reward and convenience, mostly at a cheaper price tier (§\ref{sec:rq3}). In the other direction, on 30.5\% of bound requests the always-sequential plan would have met every limit; these needless binds are 21.1\% of money binds and all space-only binds, which CP-SAT resolves inside the packing capability anyway. Treating ``always-sequential would break a limit'' as the class to detect, the gate has precision 0.78 and recall 1.00 by the validator.

\noindent\textbf{Deciding Whether to Coordinate Must Be Cheap.} On an idle machine the gated system takes 1.66 s per request at the median, against 2.04 s for always-joint and 1.85 s for validate-and-repair, and 6.9 s at the 95th percentile against 7.3 s and 8.0 s. Per accepted plan it is the cheapest method that keeps every limit on this grid (3.28 s, against 4.13 s and 3.54 s), and where nothing binds it costs 1.48 s against 1.38 s for always-sequential. This held only after one change: our first timing study found the gated system slower than validate-and-repair although it made the fewest proposals, because the gate, the trial checks and certification simulated the same plan's cost again and again. Remembering each plan's simulated totals, which returns exactly the same totals, cut its median time 1.5 times.

\noindent\textbf{With Slow Components, Proposals Decide.} Our components answer in milliseconds; a language-model or remote component takes seconds per call. With a latency of $t=1$ s per proposal (Figure~\ref{fig:costlatency}), the gated system needs 29.6 s per budgeted request, validate-and-repair 32.0 s and always-joint 51.6 s; over the core run, gating makes 54\% [53--55\%] of always-joint's proposals. The saving comes from requests where nothing binds, which propose each node once, and from bound requests whose closure is a proper subset. Against validate-and-repair the ranking depends on tightness: on the budgets users chose (median $\tau=1.51$, where some pool binds for 34\% of requests), validate-and-repair has little to repair and is slightly faster (18.1 s against 20.2 s), while always-joint takes 2.8 times as long as gating.

\noindent\textbf{The Saving Grows With Composition Size.} In the synthetic systems (§\ref{sec:setup}), the gated system finds a plan as often as always-joint, and more often at $k=20$ (80\% against 74\%), where always-joint sometimes hits the solver's time limit. It saves 25\% of always-joint's work at $k=3$ and 41\% at $k=20$ (Figure~\ref{fig:scaling}), at a utility loss of 4\% to 10\%, because capabilities decided outside a closure no longer trade preference with those inside it. At $k=20$ it needs 8\% of always-joint's time at 10 ms per component query and 49\% at 1 s. Here the validator rejects up to 8\% of the plans of the gated system, always-joint and validate-and-repair, all marginally: choosing the best of many certified plans brings the winner's curse back on the certification sample, and a 1\% margin at certification removes every rejection.

\takeaway{Whether capabilities must be solved together depends on how tight the budget is, not on the kind of request. Solving only the binding pool's closure makes the gated system the cheapest feasible method per accepted plan on our budget grid, once its own check was made cheap. Its saving lies in proposals, so it grows with slow components and more capabilities; on loose budgets, validate-and-repair is as fast.}

\subsection{RQ3: Quality}
\label{sec:rq3}
RQ3 asks how much plan quality the gate gives up. We compare each plan with the oracle's best feasible plan for the same request, both scored with one utility $U$: the itinerary reward rescaled to the trip's whole reward range and the budget plan's convenience, with equal weights. The composite solver uses the same weights but rescales the reward over its own feasible settings, so the oracle's $U$ is what puts every method on one scale. The utility gap is the oracle's $U$ minus the plan's (0 is best). We also count lost plans, where the oracle finds a plan and the method does not.

\begin{figure*}[t]
\begin{minipage}[t]{0.32\textwidth}
\centering
\includegraphics[width=\linewidth]{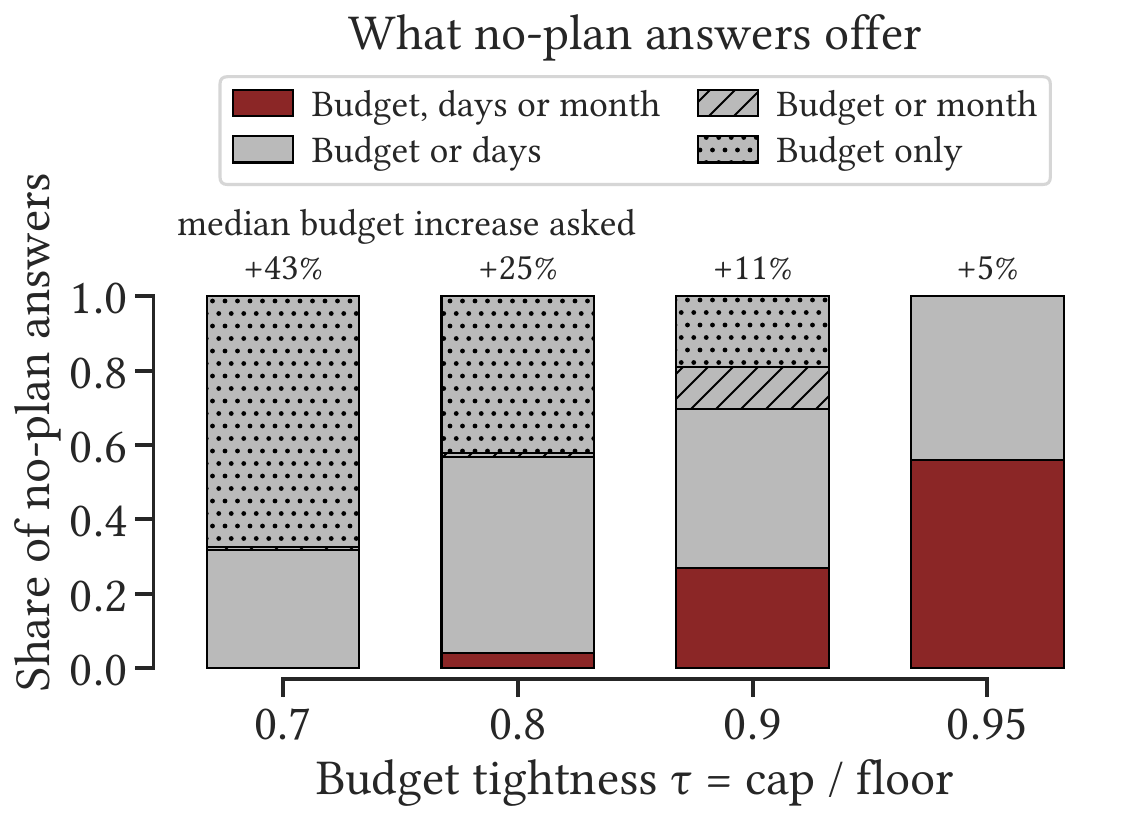}
\caption{The closer a request is to the floor, the more alternatives it is offered and the less extra budget it needs.}
\label{fig:menu}
\end{minipage}\hfill
\begin{minipage}[t]{0.32\textwidth}
\centering
\includegraphics[width=\linewidth]{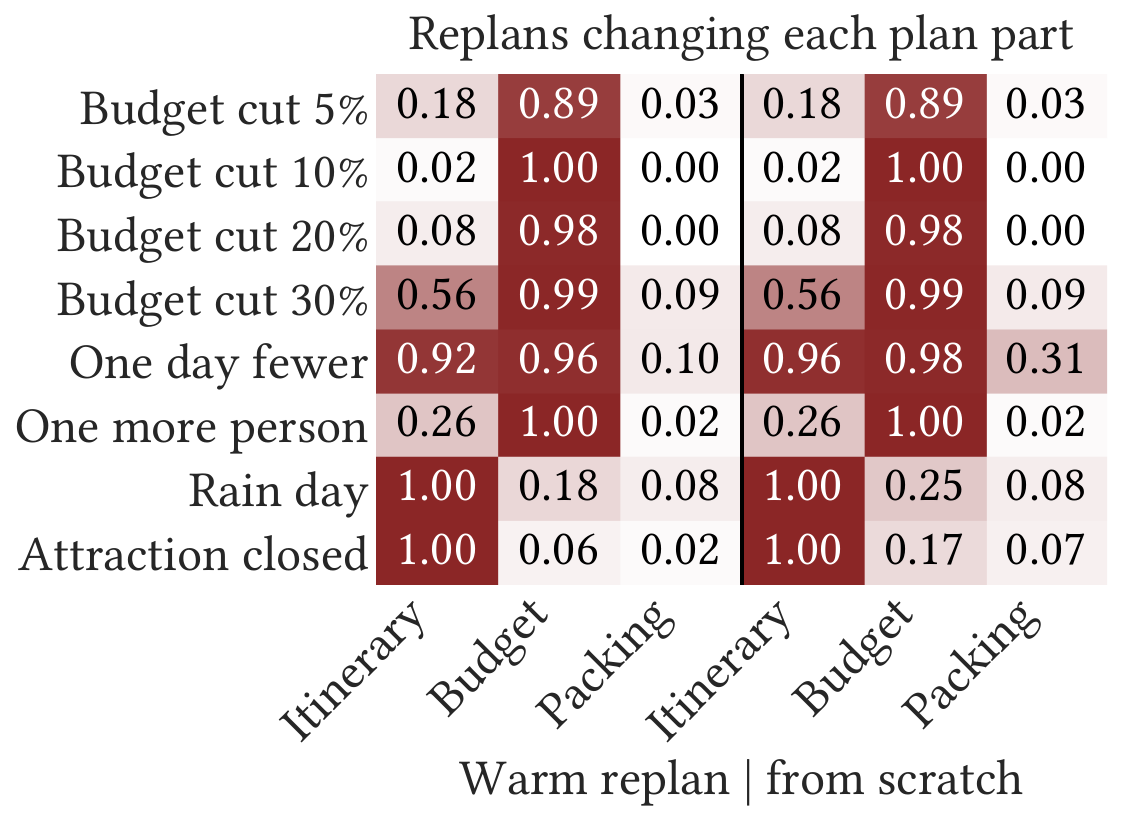}
\caption{After budget cuts, replans change budget options, rarely packing; warm repairs keep more after closures.}
\label{fig:locality}
\end{minipage}\hfill
\begin{minipage}[t]{0.32\textwidth}
\centering
\includegraphics[width=\linewidth]{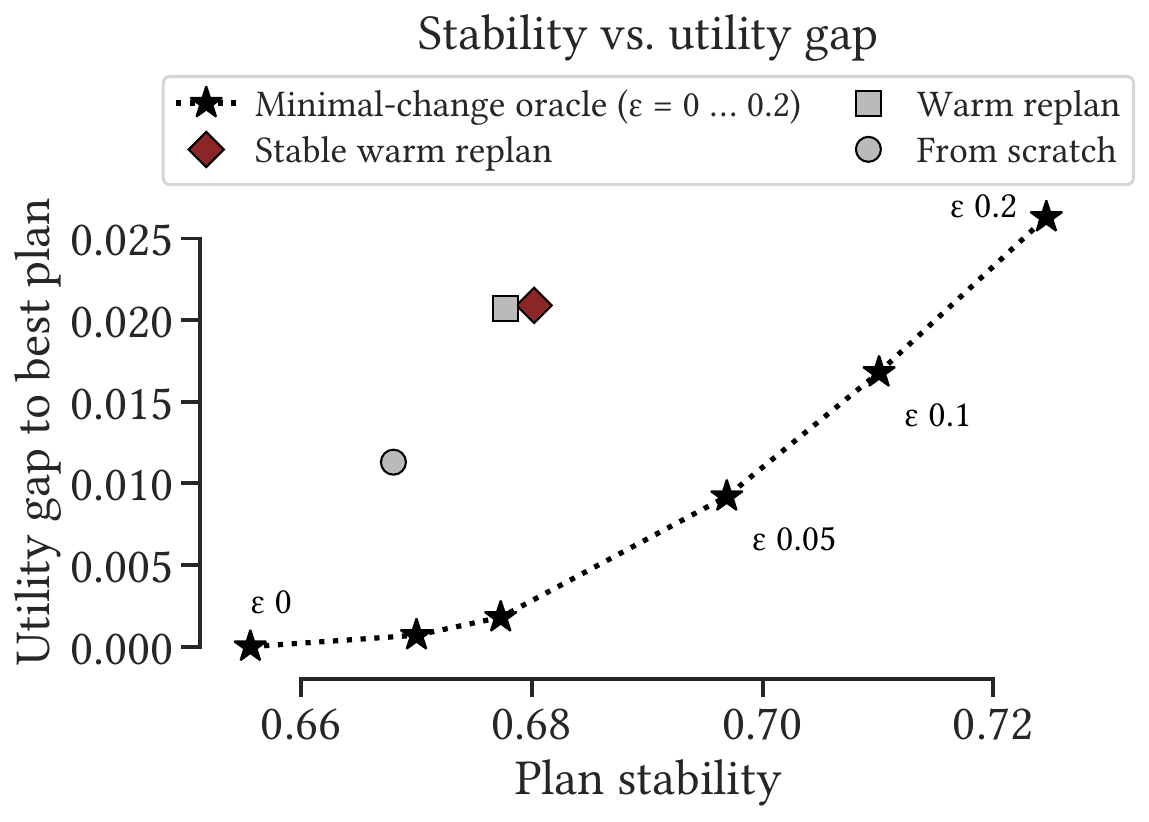}
\caption{Warm replans come close to the minimal-change oracle's stability but lose more utility.}
\label{fig:tradeoff}
\end{minipage}
\end{figure*}

\begin{table}[t]
\centering\footnotesize
\setlength{\tabcolsep}{3pt}
\caption{Quality against the exhaustive oracle over the same planners on all 2{,}728 requests; gap means are over plans both release.}
\label{tab:oracle}
\begin{tabular}{@{}lccccc@{}}
\toprule
Method & \thead{lost\\plans} & \thead{no-plan\\agreement} & \thead{gap\\mean} & \thead{gap\\$=0$} & \thead{Pareto\\frontier} \\
\midrule
Gated & 0& 1.00 & 0.0067& 92\%& 95\%\\
Always-joint & 0 & 1.00 & 0.0009& 96\%& 98\%\\
\midrule
\multicolumn{6}{@{}l}{\emph{oracle over all five planners}} \\
Gated & 0 & 1.00 & 0.168& 14\%& 16\%\\
Always-joint & 0 & 1.00 & 0.161& 15\%& 16\%\\
\bottomrule
\end{tabular}
\end{table}

\noindent\textbf{No Plan Is Lost.} Over the same planners, the gated system releases a plan for every request where the oracle finds one and answers ``no plan'' exactly where the oracle does (Table~\ref{tab:oracle}); early commitment without backtracking never cost a plan in our runs. On one request (Virginia Beach, 3 days, joint budget, $\tau=1.05$), always-sequential's plan reaches utility 0.58 but costs \$1{,}838 at the 90\% level, over the \$1{,}789 cap; the gated plan, at utility 0.19, is the best plan within the cap, the same as always-joint's and the oracle's. Independent plan checks agree: no released plan breaks an itinerary hard constraint.

\noindent\textbf{The Gap Is Small and Lies Where Money Does Not Bind.} The gated system's mean utility gap is 0.0067 [0.002--0.013], and 92\% [88--95\%] of its plans are the oracle's best (always-joint: 0.0009). Where a pool bound, 96\% of gated plans are the oracle's best; where nothing bound, 86\% are, so the gap appears only at $\tau\ge1.25$ (Figure~\ref{fig:decomp}). The misses lose itinerary reward (Figure~\ref{fig:components}): in 32 of the 89 misses where nothing bound, the oracle's best plan uses a price tier below the city's, where the planners find itineraries the reward model scores higher, a lever the gated system pulls only inside a joint solve.

\noindent\textbf{Sampling Whole Plans Needs Many More Checks.} A cheaper alternative is to sample whole combinations, check each against every limit and keep the best that fits. Drawing from the oracle's own space (Figure~\ref{fig:bestofk}), one draw finds a plan for 25\% of plannable requests, 16 draws for 72\% and 128 draws for 96\% (gated: every plan, at a median of 14 proposals). Drawing only at the city's own price tier, as decisions made one after another do, levels off at 75\%. Each draw here is checked for free; in a running system it would cost a run of every component.

\noindent\textbf{Planner Breadth Matters More Than Coordination.} Against an oracle that also uses the two slower planners, both systems trail by about 0.168: the missing planners cost at least 0.13 at every tightness, selective coordination at most 0.012 (Figure~\ref{fig:decomp}).

\takeaway{Against an exhaustive search over the same planners, the gated system finds every plan and scores 0.0067 lower on a 0--1 utility scale. The loss comes from requests whose budget does not bind, often through a cheaper price tier tried only inside a joint solve. Running more planners would gain far more than more coordination.}

\subsection{RQ4: Recovery}
\label{sec:rq4}
RQ4 asks whether the same machinery supports recovery. A relaxation is \emph{minimal} if it equals the oracle's smallest change in that dimension, and an answer's \emph{coverage} is the share of dimensions with an oracle-confirmed single change that it offers. For replanning we measure stability, the share of each part of the plan kept, next to plan quality.

\begin{figure*}[t]
\begin{minipage}[t]{0.32\textwidth}
\centering
\includegraphics[width=\linewidth]{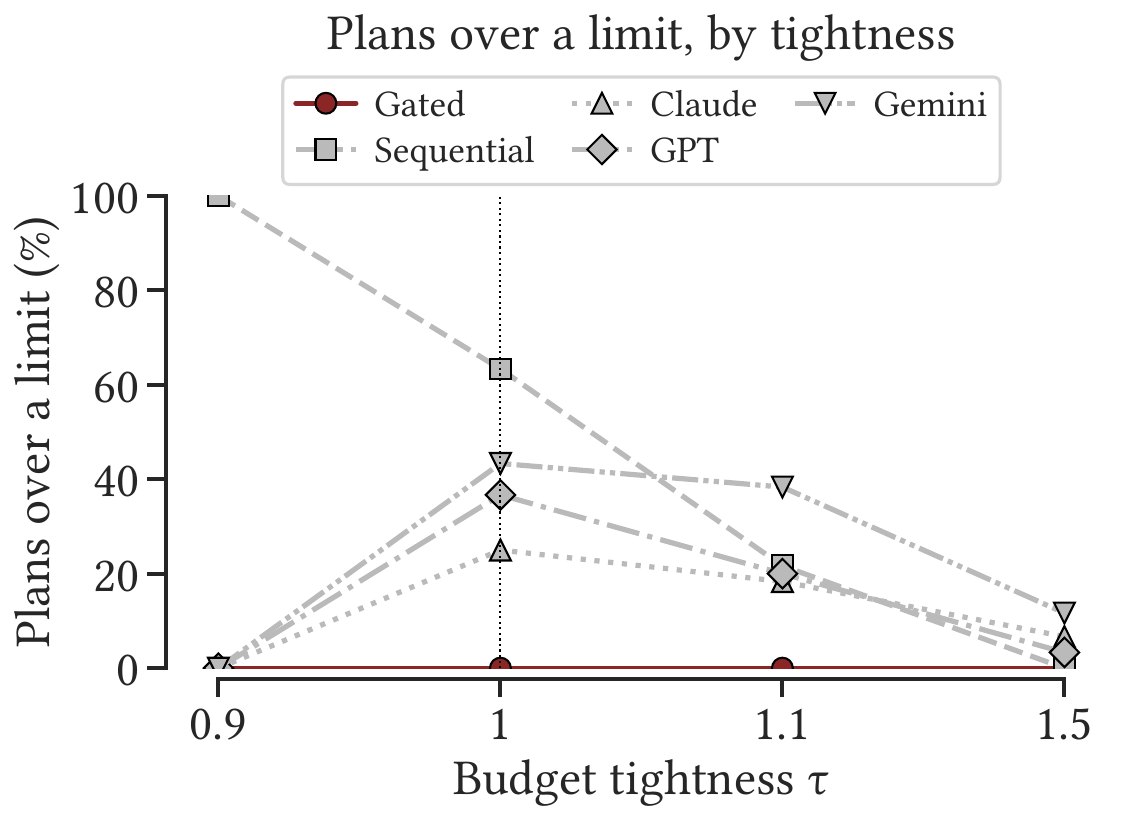}
\caption{Like always-sequential execution, agents with the verifier break a limit near the floor; gated never does.}
\label{fig:llmover}
\end{minipage}\hfill
\begin{minipage}[t]{0.32\textwidth}
\centering
\includegraphics[width=\linewidth]{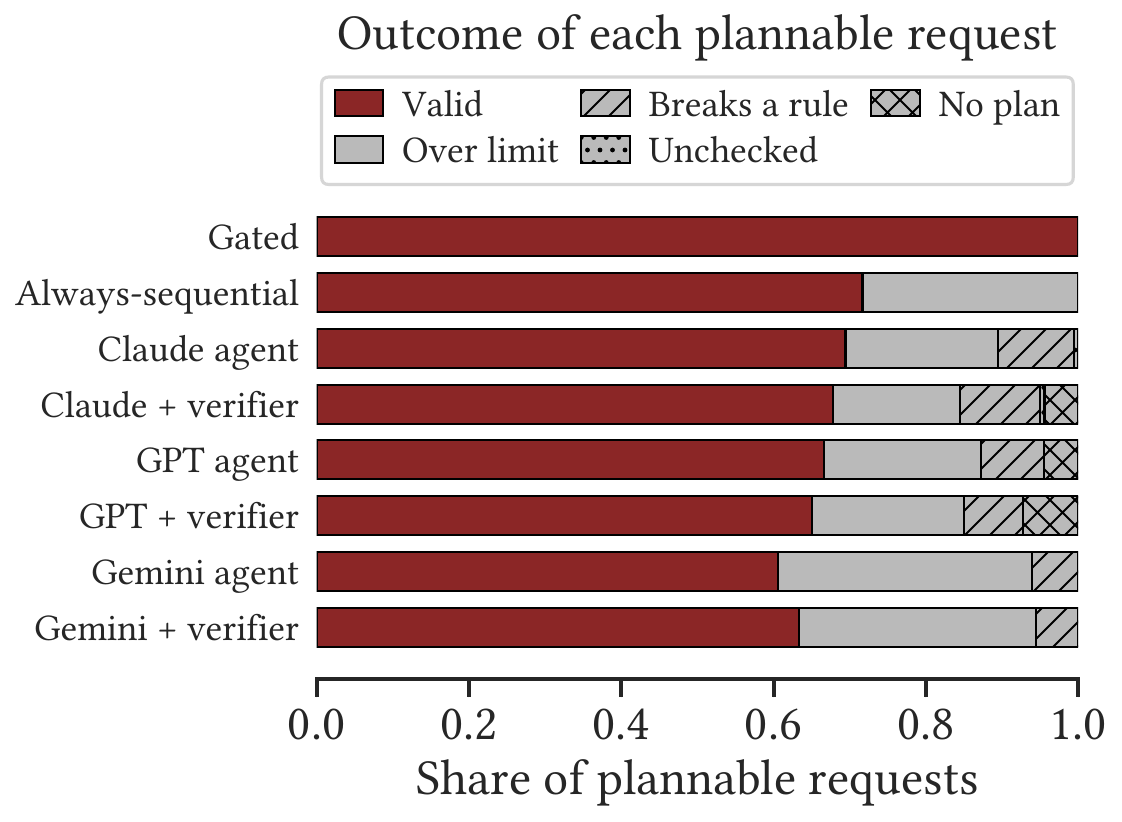}
\caption{Agents give a valid plan for 61--69\% of plannable requests; the rest break a limit or a rule.}
\label{fig:llmoutcomes}
\end{minipage}\hfill
\begin{minipage}[t]{0.32\textwidth}
\centering
\includegraphics[width=\linewidth]{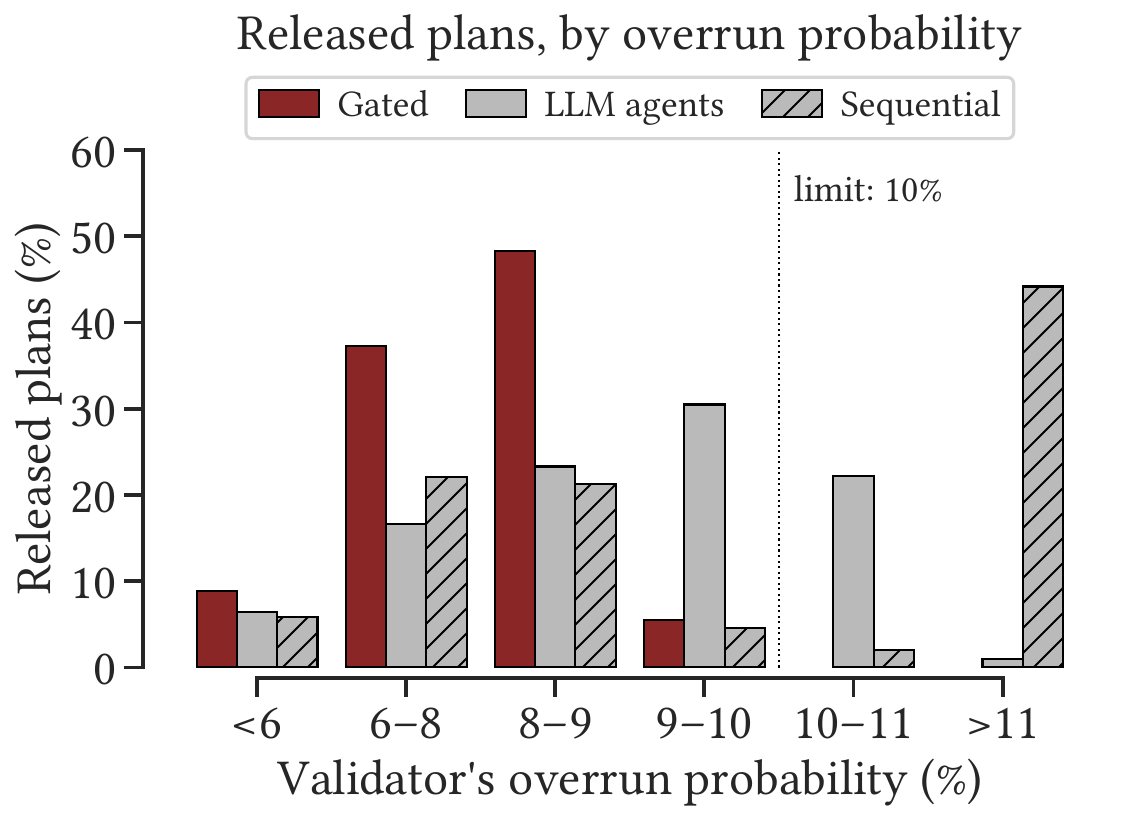}
\caption{Agents' plans pile up just under and just over the 10\% limit; certified gated plans stay below it.}
\label{fig:llmoverrun}
\end{minipage}
\end{figure*}

\noindent\textbf{Relaxations Cover Every Confirmed Change.} The old answer, ``raise the budget to about \$X'', has coverage 0.62. The new answer reaches 1.00 within the searched range: it misses no oracle-confirmed change and offers none the oracle rejects. The closer a request is to the floor, the richer the answer (Figure~\ref{fig:menu}): at $\tau=0.7$, 67\% of answers can only offer more budget, a median 43\% of the cap, while at $\tau=0.95$, 56\% offer all three dimensions and the budget increase is 5\%. Day and month changes are always minimal, and budget changes are within the search's 0.2\% tolerance for 96\% of requests. Starting the budget search from the gate's own estimate of the smallest sufficient budget cuts its median cost from 51 s to 11 s at the same coverage.

\noindent\textbf{Closures Bound Replanning.} After a budget cut, the parts that change lie outside the money closure in 2.6\% of cases (Figure~\ref{fig:locality}): budget options change in most replans, packing in at most 9\%. After an attraction closes, warm replanning keeps 0.81 of the itinerary against 0.35 from scratch. For budget cuts and added people, warm and scratch replans are identical, because the planners are deterministic and the re-opened money closure re-optimizes every decision in it. Rain is hard to absorb: with outdoor attractions excluded for a day, replans lose 1.9 visits on average, because few indoor attractions in the traveler's interests are open. Against a minimal-change oracle that keeps the most stable plan within $\varepsilon=0.05$ of the best utility (Figure~\ref{fig:tradeoff}), which keeps 0.697 of the previous plan at a utility loss of 0.009, warm replans with the stability preference keep 0.680 (0.021) and replans from scratch 0.668 (0.011). The composite solver keeps one budget plan per itinerary choice, so within $\varepsilon$ there is rarely a more stable setting to choose.

\takeaway{When no plan fits, the system offers the smallest change in budget, trip length or month that gives one, covering every change an exhaustive search confirms. Replanning touches only the parts a change reaches: a closed attraction leaves 81\% of the itinerary in place, against 35\% from scratch.}

\subsection{RQ5: Language Models}
\label{sec:rq5}
RQ5 asks whether a language model can take over either end of the system. As parsers, Claude Sonnet~5 alone and the hybrid parser read the 300 constructed requests. As orchestrators, Claude Opus~5.5, GPT-6 Astra and Gemini~4 Argon answer 260 requests of 20 test-city trips at $\tau\in\{0.9,1,1.1,1.5\}$, paired with the core run's plans. Each model runs at high reasoning effort as an \emph{agent}: it receives the typed request and the 90\% budget rule, and the three capabilities as tools (places with opening hours and visit lengths, travel times, itinerary planning and scoring, budget options with their compatibility rules, cost estimates with the chance of exceeding the cap on the gate's own 2,000 scenarios, and packing), and submits one plan, or reports that none exists, within 40 turns. As a baseline, each model also writes the whole plan in one reply, without tools (\emph{zero-shot}). With a \emph{verifier}, the LLM-Modulo pattern~\cite{kambhampati2024llmmodulo,gundawar2024robust}, each submission is checked for budget confidence on the same 2,000 scenarios, a visit on every day, agreement between the parts, packing for every planned activity and visits within the flight times, and failures are returned for up to three revisions. A plan is \emph{valid} when it is complete, names only what the catalogs hold, passes the validator, and keeps opening hours, non-overlapping visits and compatible budget options, all of which the tools show the model.

\begin{table}[t]
\centering\footnotesize
\setlength{\tabcolsep}{2.5pt}
\caption{Agents against our methods on the same 260 requests. \emph{Given}: share answered with a plan; \emph{valid}: share of the 180 plannable requests with a valid plan; \emph{money}, \emph{rule}: released plans over the money limit or breaking an itinerary rule; \emph{Pareto front}: valid plans on the oracle's frontier. Our times are from the shared core-run machine.}
\label{tab:llmorch}
\begin{tabular}{@{}lccccccc@{}}
\toprule
Method & given & valid & money & rule & \thead{Pareto\\front} & \thead{median\\s} & \thead{\$ per\\request} \\
\midrule
Gated & 0.77 & 1.00 & 0\% & 0\% & 98\% & 2.4 & -- \\
Always-joint & 0.77 & 1.00 & 0\% & 0\% & 99\% & 2.7 & -- \\
Validate-and-repair & 0.77 & 1.00 & 0\% & 0\% & 97\% & 1.5 & -- \\
Always-sequential & 1.00 & 0.72 & 43\% & 0\% & 94\% & 0.7 & -- \\
\midrule
Claude agent & 0.78 & 0.69 & 18\% & 12\% & 76\% & 77 & 0.46 \\
Claude + verifier & 0.75 & 0.68 & 15\% & 13\% & 62\% & 92 & 0.53 \\
GPT agent & 0.75 & 0.67 & 19\% & 12\% & 56\% & 84 & 0.70 \\
GPT + verifier & 0.73 & 0.65 & 19\% & 12\% & 57\% & 84 & 0.71 \\
Gemini agent & 0.78 & 0.61 & 30\% & 8\% & 92\% & 278 & 0.42 \\
Gemini + verifier & 0.78 & 0.63 & 28\% & 8\% & 77\% & 352 & 0.53 \\
\bottomrule
\end{tabular}
\end{table}

\noindent\textbf{A Hybrid Parser Reads Requests; a Model Alone Does Not.} The hybrid parser reaches end-to-end success 0.97, with all fields right for 0.96 and no fabricated value, against 0.69 for the pattern parser, whose weaknesses with relative dates and free-form wording it removes. The language model alone does no better than the pattern parser (0.70): it misreads which outputs are wanted in 40\% of requests and plans on a guessed value in 15\% of those that need a question, both of which the hybrid's cross-check prevents. Follow-ups parse exactly in at least 99.9\% of the 17{,}072 templated messages, an upper bound since the templates were written with the pattern parser, and all 3{,}344 explanations carry only numbers the result holds (100\%).

\noindent\textbf{Agents Break the Shared Limits Near the Floor.} Agents release plans over the money limit in 15--30\% of cases and break an itinerary rule they were shown in another 8--13\%, against none for the gated system, always-joint and validate-and-repair (Table~\ref{tab:llmorch}). Like always-sequential execution, they fail most near the floor (Figure~\ref{fig:llmover}): with the verifier, 25--43\% of requests at $\tau=1$ receive a plan over a limit (always-sequential: 63\%), and 3--12\% at $\tau=1.5$; below the floor they correctly answer ``no plan'' for 95\% of requests. Of the 180 requests the oracle can plan, agents give a valid plan for 61--69\%, fewer than always-sequential (72\%; Figure~\ref{fig:llmoutcomes}). Zero-shot, every model writes a plan for at least 259 of the 260 requests, but none can be checked: each names places or budget options the catalogs do not hold.

\noindent\textbf{A Verifier on the Same Sample Repeats the Winner's Curse.} The verifier lowers money failures by at most 2 points. It checks on the 2,000 scenarios the agent's cost tool used, so a plan tuned to sit just under the limit on that sample passes and then overruns on a fresh one (Figure~\ref{fig:llmoverrun}): 54\% of the agents' plans have a true overrun probability between 9 and 11\%, and 96\% of those over the limit are within one point of it (median 10.3\%), while certified gated plans stay at or below 9.4\%. This is the selection effect of §\ref{sec:rq1}.

\noindent\textbf{A Fresh-Sample Test Removes the Overruns, and Most Plans With Them.} Giving the agents the gated system's release test removes their money failures, but not the rest. Re-checking every agent plan on the gated system's certification sample, and counting a failure as no plan, 54--67\% of the agents' budgeted plans fail, including 47--66 per agent that the validator accepts, because a plan tuned to sit just under the limit leaves no room for the test's confidence bound. The agents would then give a valid plan for 32--41\% of plannable requests, and 6--10\% of their remaining plans still break an itinerary rule. Every gated plan passes the same test.

\noindent\textbf{Agents Are Slow.} Agents take a median of 77--352 s and \$0.42--0.71 in API fees per request, against 2.4 s for the gated system, one to two orders of magnitude more. Their valid plans lie on the oracle's Pareto frontier in 56--92\% of cases (gated: 98\%); Gemini's itineraries score above the gated plans by the reward model, because an agent can assemble visits outside the planners' candidates.

\takeaway{Language belongs at the edges of the system: a hybrid parser reads requests almost perfectly without inventing values. Given the same capabilities and budget rule, language-model agents release plans over the money limit in 15--30\% of cases, just over it, where a verifier that checks on the sample a plan was tuned on cannot see them, and a fresh-sample test that removes these overruns removes most of their plans with them.}

\section{Related Work}
\label{sec:related}

\noindent\textbf{Agents, Tools and Verifiers.} Language agents interleave reasoning with tool calls~\cite{yao2023react,schick2023toolformer} or orchestrate specialized models~\cite{shen2023hugginggpt,wu2024autogen}, and agentic recommenders learn tool chains~\cite{li2026chainrec} or plan over conventional recommenders~\cite{yang2026prta}. These systems choose which tools to call and in what order, but none tracks resources the tools share. TravelPlanner shows that language agents fail hard constraints in travel planning~\cite{xie2024travelplanner}. LLM-Modulo frameworks pair generation with external verifiers~\cite{kambhampati2024llmmodulo,gundawar2024robust}, and solvers can guarantee feasibility when a model translates the request into one formal problem~\cite{hao2025formal,ju-etal-2024-globe,liu2026hard}. We share the separation of language from enforcement, but compose existing feasibility-preserving capabilities rather than one formulation, and decide at run time which of them must be solved together.

\noindent\textbf{Contracts, Budgets and Decomposition.} Typed pre- and postcondition interfaces have precedents in contract-based design and semantic service composition~\cite{benveniste2018contracts,martin2004owls}. Budget control in multi-agent systems delegates or allocates budgets top-down: Agent Contracts add conservation laws to delegated budgets~\cite{ye2026agent}, and ZEBRA splits a budget across pipeline phases by estimated utility~\cite{hamri2026zebra}. Price-based decomposition coordinates subproblems through a dual price on the shared limit~\cite{palomar2006decomposition}, but needs components that re-optimize against a price and many rounds of proposals. In our setting, existing capabilities claim one uncertain pool and no parent allocates it first.

\noindent\textbf{Escrow and Treaties.} The closest database precedent maintains a shared numeric constraint across parties that act separately. The escrow method reserves quantities of an aggregate so that transactions commit without blocking~\cite{oneil1986escrow}, the demarcation protocol gives each site a local limit and coordinates only when a site must move it~\cite{barbara1994demarcation}, and the homeostasis protocol derives such treaties from transaction code~\cite{roy2015homeostasis}; coordination avoidance asks when an invariant can be kept without coordinating at all~\cite{bailis2014coordination}. Our gate follows the same logic, acting alone while slack remains, with two differences: the claims are uncertain and correlated, so slack is a quantile of a joint total rather than a sum of reserved amounts, and the parties are optimizers whose decisions can be re-chosen jointly, not transactions that only commit or wait.

\noindent\textbf{Uncertain Data and Package Queries.} Package queries apply constraints to a set of rows collectively~\cite{10.14778/2904483.2904489}, and stochastic package queries add uncertain attributes and probabilistic constraints, validating packages on scenarios separate from those used to choose them~\cite{10.1145/3318464.3389765,10.14778/3746405.3746431}, as our certification does. MCDB evaluates queries over Monte Carlo instances of uncertain data~\cite{jampani2008mcdb}, and robust optimizers choose plans by a quantile of their cost distribution~\cite{babcock2005robust}. These systems optimize one query or package over one relation, not a composition of separately owned optimizers.

\noindent\textbf{Chance Constraints and Recovery.} Chance-constrained programming~\cite{charnes1959chance,nemirovski2006convex} underlies our money pool, used here as a run-time test of whether decomposition keeps the limits. Its sample-based forms validate a solution on scenarios not used to find it~\cite{luedtke2008sample}, the remedy for the optimizer's curse~\cite{smith2006optimizer}. Conflict explanation and relaxation have a long history in constraint satisfaction and recommendation~\cite{junker2004quickxplain,felfernig2015constraint}, and plan stability has been studied as replanning versus repair~\cite{fox2006planstability}; our relaxations search one dimension at a time under a probabilistic budget and confirm each by running the system, and closures bound how far a replan propagates.

\section{Discussion}
\label{sec:discussion}

\noindent\textbf{Lessons.} The four lessons apply wherever separately built components share an uncertain resource. The joint cost model is the one piece no component can supply, so it belongs to the orchestrator. Certification must be kept apart from selection at every point where candidates are compared, including after a failed test, as the synthetic benchmark and the agents' verifier both show. Certification cannot correct the cost model, so realized costs, where they exist, should recalibrate it rather than widen the margin. And gating saves proposals, not computation inside them: it pays most near the floor, with slow components and with many capabilities sharing a pool, while on loose budgets validate-and-repair is as fast.

\noindent\textbf{Limitations.} The requests are built from real trips and catalogs but not collected from users. No realized cost exists for any released plan, so model error is tested only by perturbing the model. A plan chosen after a failed certification is selected on the certification sample; none failed the validator, but a second fresh sample would restore the exact bound. Cross-capability levers such as the price tier are tried only inside a joint solve, the composite solver keeps one budget plan per itinerary and never backtracks, and the system runs three of five planners, which costs more utility than coordination does. Times come from one idle machine, and the five-planner oracle covers 20 of the 88 trips. The language-model orchestrators ran once on 260 requests, and the fresh-sample test was applied afterwards, without letting an agent revise a rejected plan. A human study of plan and explanation quality is not yet part of the evaluation, and the evaluation covers one domain, extended only synthetically.

\section{Conclusion}
\label{sec:conclusion}
Local feasibility is not preserved under composition: running three individually sound travel capabilities one after another broke a shared limit for 51\% of our requests. Joint solving is needed only when a shared pool is tight, and then only for that pool's resource closure. An orchestrator that predicts binding at run time from one joint cost model and certifies every release on scenarios independent of the gate's sample released no plan an independent re-simulation rejects and missed no plan an exhaustive search over the same planners finds, at about half the proposals of always solving jointly. The same machinery turns ``no plan'' into confirmed, minimal relaxations and keeps replanning inside the closures a change reaches. Feasibility for composed systems rests on a margin computed on the joint total, on certification kept apart from selection, and on a cost model that is itself checked. Language models belong at the edges of such a system: they read requests well, but orchestrating the same capabilities they break the shared limits the gate keeps.

\section*{GenAI Usage Disclosure}
During the preparation of this work, the authors used LLM-based tools to assist with condensing and improving the clarity of the author-written text within selected sections. AI tools were not used to generate any original content, results, analyses, or claims. All AI-assisted edits were reviewed, verified and approved by the authors. The authors take full responsibility for the accuracy and integrity of all content in this paper.

\bibliographystyle{ACM-Reference-Format}
\bibliography{references}

\end{document}